\documentclass{article}

\usepackage[preprint]{neurips_2026}

\usepackage[utf8]{inputenc}
\usepackage[T1]{fontenc}

\usepackage{amsmath}
\usepackage{amssymb}
\usepackage{amsfonts}

\usepackage{booktabs}
\usepackage{multirow}
\usepackage{colortbl}
\usepackage[table,xcdraw]{xcolor}  
\usepackage{arydshln}
\usepackage{rotating}
\usepackage{diagbox}
\usepackage{tcolorbox}
\tcbuselibrary{skins,breakable}
\usepackage{multicol}
\usepackage{capt-of}
\usepackage{needspace}
\usepackage{placeins}

\usepackage{graphicx}
\usepackage{caption}
\usepackage{subcaption}
\usepackage{rotating}
\usepackage{adjustbox}

\usepackage{nicefrac}
\usepackage{microtype}
\usepackage{xfp}
\usepackage{pifont}

\usepackage{hyperref}
\usepackage{url}

\title{Enhancing Multimodal In-Context Learning via Inductive-Deductive Reasoning}

\author{
  \textbf{Haoyu Wang}$^{1,2\ast}$, 
  \textbf{Haonan Wang}$^{2\ast}$, 
  \textbf{Yuyan Chen}$^{3}$, 
  \textbf{Jun Chen}$^{1}$,\\
  \textbf{Gang Liu}$^{1}$, 
  \textbf{Qian Wang}$^{1}$$^{\dag}$, 
  \textbf{Jiahong Yan}$^{1}$, 
  \textbf{Yanghua Xiao}$^{2}$$^{\dag}$ \\
  \\ 
  $^{1}$Tencent QQ \quad
  $^{2}$Fudan University \quad
  $^{3}$Cornell University \\
  \texttt{\{wanghy24@m., wanghn24@m., shawyh@\}fudan.edu.cn} \\ 
  \texttt{yolandachen0313@gmail.com} \\
  \texttt{\{albertjchen, sinbadliu, alongwang, redyan\}@tencent.com}
}

\author{
  \textbf{Haoyu Wang}$^{1,2\ast}$, 
  \textbf{Haonan Wang}$^{2\ast}$, 
  \textbf{Yuyan Chen}$^{3}$, 
  \textbf{Jun Chen}$^{1}$,\\
  \textbf{Gang Liu}$^{1}$, 
  \textbf{Qian Wang}$^{1}$$^{\dag}$, 
  \textbf{Jiahong Yan}$^{1}$, 
  \textbf{Yanghua Xiao}$^{2}$$^{\dag}$ \\
  \\ 
  $^{1}$Tencent QQ \quad
  $^{2}$Fudan University \quad
  $^{3}$Cornell University \\
  \texttt{\{wanghy24, wanghn24\}@m.fudan.edu.cn} \\ 
}

\begin{document}

\maketitle

\hypersetup{hidelinks}
\renewcommand{\thefootnote}{}
\footnote{Work done during internship at Tencent QQ. $^{\ast}$ Equal contribution. $^{\dag}$ Corresponding author.}


\begin{abstract}
    In-context learning (ICL) allows large models to adapt to tasks using a few examples, yet its extension to vision-language models (VLMs) remains fragile. Our analysis reveals that the fundamental limitation lies in an inductive gap — models often produce correct answers from flawed reasoning, while struggling to extract consistent rules across demonstrations. This gap is further exacerbated by two visual-level obstacles: an overwhelming proportion of redundant visual tokens that obscure textual cues, and a skewed attention distribution that favors the initial image at the expense of subsequent context. To address these issues, we introduce a framework that restructures multimodal ICL as a principled inductive-deductive process. The framework incorporates a similarity-based visual token compression module to filter out redundant patches, a dynamic attention rebalancing mechanism to distribute focus equitably across all images, and a chain-of-thought paradigm that explicitly guides the model to analyze individual examples, derive a generalizable rule, and then apply it to the query. An auxiliary learning pipeline combines supervised fine-tuning with reinforcement learning using verifiable rewards to reinforce faithful citation and noise filtering. Evaluations across eight benchmarks covering visual perception, logical reasoning, STEM problems, and sarcasm detection demonstrate consistent and significant improvements over standard ICL baselines for multiple open-source VLMs, highlighting the potential of equipping models with genuine inductive capabilities in multimodal settings.
\end{abstract}

\section{Introduction}

\begin{figure}[t]
    \centering
    \includegraphics[width=\textwidth]{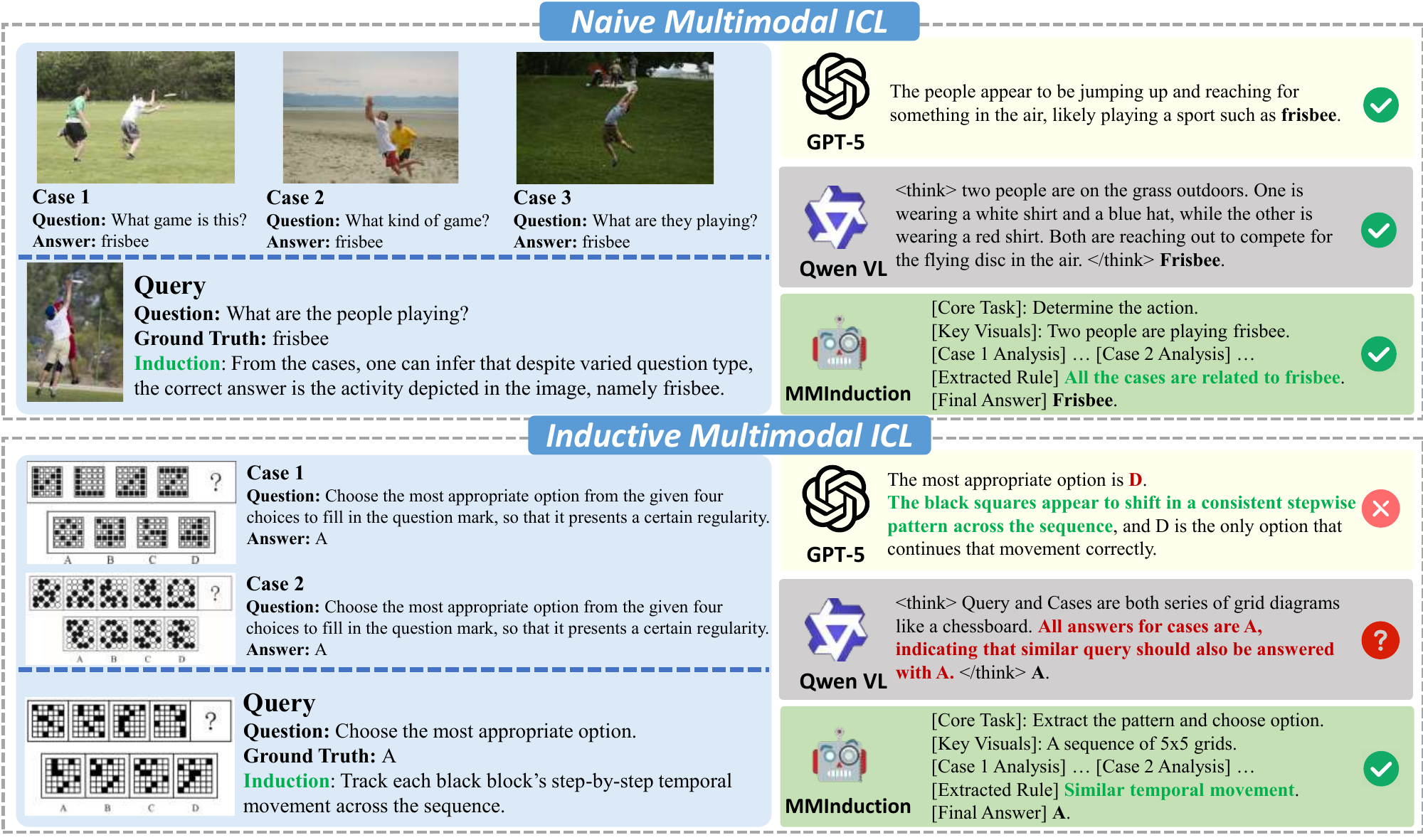}
    \caption{Examples of naive and inductive multimodal ICL. For visual perception-based QA, VLMs quickly grasp the task through cases and derive answers based on similar visual elements. However, for complex reasoning, models may either produce incorrect answers by incomplete patterns, or fail to generalize commonalities from the cases, leading to the failure of ICL.}
    \label{fig:fig1}
    \vspace{-2em}
\end{figure}

In-Context Learning (ICL) \cite{dong2024survey} has emerged as a key capability of large language models (LLMs), enabling them to improve task performance by leveraging a few demonstrations provided in the prompt \cite{wei2021finetuned,wies2023learnability,min2022rethinking}. At its core, ICL can be viewed as an induction-deduction process: models first induce the underlying patterns from a set of reference examples (i.e., \textit{cases}) and then apply these patterns to solve the target problem (i.e., \textit{query}) \cite{cai2025role,cheng2024inductive}. Given the success in text-only settings, a natural question is whether similar benefits extend to multimodal scenarios. However, in the domain of multimodality, represented by Vision-Language Models (VLMs) \cite{liu2023visual}, the potential of multimodal ICL remains largely underexplored \cite{baldassini2024makes,chen2025can,doveh2024towards}. 

As illustrated in Figure \ref{fig:fig1}, while VLMs adequately handle tasks that rely on surface-level visual similarity, when it comes to complex reasoning, they either produce answers through flawed thinking or fail to generalize commonalities from the cases. To isolate the root cause of this failure, we conduct a diagnostic evaluation on the graphic reasoning benchmark MMIQ \cite{cai2025mm}, where we explicitly judge the models’ reasoning chains to determine whether the induced pattern from the cases is correct. As shown in Table \ref{tab:induction}, the accuracy to induce correct patterns is lower than task accuracy. A substantial portion of correct answers are derived from incorrect or incomplete induction. For instance, in the 1-shot setting, only 2.70\%  of Qwen3VL-8B’s \cite{bai2025qwen3} outputs are based on correct inductive reasoning, while 25.23\%  reach the right answer through erroneous logic. In the majority of these failures, the generated reasoning chains barely reference the provided cases even when expressly prompted to reference them, indicating that models tend to rely on pre-trained priors rather than on the in-context demonstrations. This mismatch exposes a critical \textit{inductive correctness gap}, where models can luckily guess the correct answer without reliably inducing the underlying rule.  We thus identify this inductive gap as a core bottleneck of multimodal ICL.

To overcome this fundamental inductive gap, we propose \textbf{MMInduction}, a framework that instructs multimodal ICL as a structured inductive-deductive process. At its core lies an Inductive Chain-of-Thought (CoT) \cite{wei2022chain}, a reasoning template that guides the model to first analyze the provided cases, extract a generalizable rule, and then apply that rule to deduce the answer for the query. This stands in stark contrast to the naive jump-to-answer pattern prevalent in existing multimodal ICL. To equip the model with robust inductive-deductive reasoning capabilities, we adopt a combined training paradigm of Supervised Fine-Tuning (SFT) \cite{hu2022lora} followed by Reinforcement Learning with Verifiable Rewards (RLVR) \cite{shao2024deepseekmath}. The SFT stage uses high-quality reasoning chains distilled from a strong teacher model to cold-start the structured reasoning format. The RLVR stage then employs a carefully designed composite reward function that not only evaluates final answer correctness but also rewards the model for identifying irrelevant "noise" cases and correctly citing helpful ones. This fine-grained supervision enriches the training signal and enhances the model's robustness against misleading context.

However, merely imposing this reasoning structure is insufficient. Our motivation study reveals that the induction is sabotaged by two visual representation barriers. First, the overwhelming volume of visual tokens from multiple images drowns out the textual evidence essential for cross-case rule extraction. Second, the spontaneous visual attention bias towards the first image prevents the model from fully utilizing information from cases and the query. To create a clean and balanced visual foundation, we introduce two lightweight support modules: Similarity-Guided Visual Token Pruning, which compresses redundant visual inputs while preserving semantic integrity, and Dynamic Visual Attention, which rebalances the model's attention distribution across all context images during decoding. Together, they provide a necessary foundation for the higher-level reasoning process.

Extensive experiments on eight datasets spanning visual perception, logical reasoning, STEM question answering, and sarcasm detection demonstrate that MMInduction significantly improves the ICL capabilities of three open‑source VLMs. Our contributions can be summarized as follows: (1) We identify and empirically validate that the core deficiency in multimodal ICL is a lack of genuine inductive reasoning ability, systematically accompanied by the representation-level barriers of visual token redundancy and attention imbalance. (2) We present MMInduction, a novel inductive-deductive paradigm for multimodal ICL that integrates structured reasoning with targeted visual processing modules. (3) Comprehensive evaluations verify the effectiveness of our framework and reveal the broader potential of multimodal in-context learning.

\section{Related Work}
\label{sec:related_work}

\noindent \textbf{Multimodal In-Context Learning.} Existing research on multimodal ICL primarily focuses on optimizing the structure of VLM, including multi-case visual feature fusion \cite{li2024implicit}, visual attention control \cite{chen2025true}, and test-time intervention \cite{tian2025identifying}. M$^{2}$IV \cite{li2025m} directly injects a set of learnable multimodal context vectors into the context of the VLM. DARA \cite{chen2025true} proposes a parameter-efficient fine-tuning approach that dynamically reallocates attention scores to visual tokens. CAMA \cite{li2026make} designs two-stage attention modulation based on the input context sequence, thereby enhancing the focus on semantically important tokens. CATP \cite{li2026catp} introduces a training-free, progressive token pruning strategy based on cross-modal alignment. However, these methods still lack an explicit induction–deduction process, and most of their downstream tasks focus on vision-centric QA. MMInduction explicitly empowers the model to extract underlying rules from cases before applying them, thereby elevating Multimodal ICL from shallow pattern matching to real inductive reasoning.

\noindent \textbf{Reasoning in Multimodal LLMs.} The rapid progress of large language models, such as DeepSeek R1 \cite{guo2025deepseek} and Kimi \cite{team2026kimi}, has fueled growing interest in complex reasoning tasks, while multimodal large language models aim to extend this capability into the visual domain for cross-modal applications \cite{thawakar2025llamav,yang2025r1,zhao2025r1}.  Recent VLMs further improve visual reasoning through techniques like visual Chain-of-Thought \cite{zhang2023multimodal}, enhanced search strategies \cite{wu2025mmsearch,yao2026mm}, reasoning data organization \cite{wang2026mmifevol,huang2025vision}, agentic reinforcement learning \cite{liu2026infiguiagent,hong2025deepeyesv2}, and image-based reasoning trajectories \cite{su2025thinking,zheng2025deepeyes}, contributing to advances in both performance and interpretability.

\section{Motivation Study}
\label{sec:motivation}

Prior work on multimodal ICL largely focuses on visual perception tasks (e.g., VQA \cite{antol2015vqa}, OCR \cite{yang2025cc}), where correct answers can often be inferred from surface-level visual similarity without discovering complex patterns across cases. The genuine ICL capabilities of VLMs on reasoning-intensive tasks remain largely unexplored. Therefore, we select three categories of benchmark : VQA \cite{antol2015vqa}, MMIQ \cite{cai2025mm}, and MDK12 \cite{zhou2026mdk12}, to examine visual perception, logical reasoning, and STEM knowledge, respectively. We partition a subset of data as a knowledge base and retrieve the Top-$k$ most similar knowledge base samples for the remaining evaluation data, injecting them as cases into the prompt context. Evaluation settings are detailed in Appendix, results are illustrated in Table~\ref{tab:accuracy}.


\begin{table}[htbp]
\centering
\caption{Accuracy ($\%$) of VLMs under various few-shot settings across visual perception (VQA), logical reasoning (MMIQ), and STEM knowledge (MDK12) benchmarks.}
\label{tab:accuracy}

\definecolor{vqa0}{HTML}{E3EDFA}
\definecolor{vqa1}{HTML}{CADCF6}
\definecolor{vqa2}{HTML}{B1CBF1}
\definecolor{vqa3}{HTML}{98BAED}

\definecolor{mmiq0}{HTML}{DFF2EC}
\definecolor{mmiq1}{HTML}{C2E6DA}
\definecolor{mmiq2}{HTML}{A5DAC8}
\definecolor{mmiq3}{HTML}{88CEB6}

\definecolor{mdk0}{HTML}{EDE3FA}
\definecolor{mdk1}{HTML}{DCCAF6}
\definecolor{mdk2}{HTML}{CBB1F1}
\definecolor{mdk3}{HTML}{BA98ED}

\resizebox{\textwidth}{!}{%
\begin{tabular}{l
  |>{\columncolor{vqa0}}c>{\columncolor{vqa1}}c>{\columncolor{vqa2}}c>{\columncolor{vqa3}}c
  |>{\columncolor{mmiq0}}c>{\columncolor{mmiq1}}c>{\columncolor{mmiq2}}c>{\columncolor{mmiq3}}c
  |>{\columncolor{mdk0}}c>{\columncolor{mdk1}}c>{\columncolor{mdk2}}c>{\columncolor{mdk3}}c}
\toprule
\rowcolor{white}
& \multicolumn{4}{c|}{\textbf{VQA}}
& \multicolumn{4}{c|}{\textbf{MMIQ}}
& \multicolumn{4}{c}{\textbf{MDK12}} \\
\rowcolor{white}
\textbf{Model}
  & \textbf{0-shot} & \textbf{1-shot} & \textbf{2-shot} & \textbf{3-shot}
  & \textbf{0-shot} & \textbf{1-shot} & \textbf{2-shot} & \textbf{3-shot}
  & \textbf{0-shot} & \textbf{1-shot} & \textbf{2-shot} & \textbf{3-shot} \\
\midrule
GPT-5.4
  & 83.76          & \textbf{87.75} & 87.46          & 88.46
  & \underline{37.38} & \underline{37.12} & \underline{37.98} & 35.66
  & 57.53          & 63.49          & 63.54          & 64.13          \\
Gemini-3.1-pro
  & 81.77          & 86.18          & 85.04          & 85.75
  & \textbf{42.40} & \textbf{45.61} & \textbf{44.04} & \textbf{46.25}
  & \textbf{64.41} & \textbf{69.69} & \textbf{71.43} & \textbf{71.92} \\
Claude-4.6-Sonnet
  & 85.33 & \underline{87.61} & 87.89 & 89.60
  & 30.43          & 32.99          & 32.94          & 33.29
  & \underline{62.22} & 64.64          & 64.78          & \underline{66.44} \\
Qwen3.5-397B-A17B
  & \underline{85.83} & 87.18          & \textbf{88.60} & \textbf{89.89}
  & 36.26          & 35.19          & 35.08          & \underline{37.04}
  & 58.80          & \underline{66.45} & \underline{66.82} & \underline{69.30} \\
Qwen3VL-8B
  & \textbf{86.04} & \textbf{87.75} & \underline{88.18} & \underline{89.74}
  & 32.30          & 28.70          & 27.96          & 27.22
  & 46.59          & 48.47          & 48.11          & 48.98          \\
Qwen2VL-7B
  & 66.92          & 69.54          & 70.48          & 71.64
  & 26.56          & 25.50          & 26.82          & 27.45
  & 28.96          & 33.99          & 34.69          & 34.33          \\
\bottomrule
\end{tabular}%
}
\end{table}

\begin{figure*}[t]
\centering

\definecolor{ans0}{HTML}{E3EDFA}
\definecolor{ans1}{HTML}{CADCF6}
\definecolor{ans2}{HTML}{B1CBF1}
\definecolor{ans3}{HTML}{98BAED}
\definecolor{ind0}{HTML}{FFF3E0}
\definecolor{ind1}{HTML}{FFE0B2}
\definecolor{ind2}{HTML}{FFCC80}
\definecolor{ind3}{HTML}{FFB74D}
\definecolor{conf0}{HTML}{FCE4EC}
\definecolor{conf1}{HTML}{F8BBD0}
\definecolor{conf2}{HTML}{F48FB1}
\definecolor{conf3}{HTML}{F06292}

\begin{minipage}{\textwidth}
\centering
\captionof{table}{Performance (\%) of VLMs under various few-shot settings. $\checkmark$ and $\times$ denote the relationship between induction and answer (e.g., $\checkmark \rightarrow \times$ means correct induction led to wrong answer).}
\label{tab:induction}
\resizebox{0.85\textwidth}{!}{%
\begin{tabular}{ll|c|c|cccc}
\toprule
\rowcolor{white}
& & \textbf{Answer Acc.} & \textbf{Induction Acc.}
& $\checkmark\rightarrow\checkmark$
& $\checkmark\rightarrow\times$
& $\times\rightarrow\checkmark$
& $\times\rightarrow\times$ \\
\midrule
\multirow{4}{*}{\textbf{GPT-5.4}}
& 0-shot & \cellcolor{ans0} 34.35 & \cellcolor{ind0} 23.09 & \cellcolor{conf0} 18.02 & \cellcolor{conf0} 5.07 & \cellcolor{conf0} 16.33 & \cellcolor{conf0} 60.59 \\
& 1-shot & \cellcolor{ans1} 43.13 & \cellcolor{ind1} 24.89 & \cellcolor{conf1} 23.42 & \cellcolor{conf1} 1.46 & \cellcolor{conf1} 19.71 & \cellcolor{conf1} 55.41 \\
& 2-shot & \cellcolor{ans2} 40.20 & \cellcolor{ind2} 26.58 & \cellcolor{conf2} 25.11 & \cellcolor{conf2} 1.46 & \cellcolor{conf2} 15.09 & \cellcolor{conf2} 58.33 \\
& 3-shot & \cellcolor{ans3} 37.27 & \cellcolor{ind3} 25.58 & \cellcolor{conf3} 24.44 & \cellcolor{conf3} 1.13 & \cellcolor{conf3} 12.84 & \cellcolor{conf3} 61.60 \\
\midrule
\multirow{4}{*}{\textbf{Gemini-3.1-pro}}
& 0-shot & \cellcolor{ans0} \textbf{61.23} & \cellcolor{ind0} \textbf{44.30} & \cellcolor{conf0} 43.15 & \cellcolor{conf0} 1.15 & \cellcolor{conf0} 18.17 & \cellcolor{conf0} 37.53 \\
& 1-shot & \cellcolor{ans1} \textbf{58.51} & \cellcolor{ind1} \textbf{40.32} & \cellcolor{conf1} 38.91 & \cellcolor{conf1} 1.41 & \cellcolor{conf1} 19.61 & \cellcolor{conf1} 40.07 \\
& 2-shot & \cellcolor{ans2} \textbf{57.77} & \cellcolor{ind2} \textbf{42.58} & \cellcolor{conf2} 41.19 & \cellcolor{conf2} 1.40 & \cellcolor{conf2} 16.58 & \cellcolor{conf2} 40.84 \\
& 3-shot & \cellcolor{ans3} \textbf{58.31} & \cellcolor{ind3} \textbf{40.34} & \cellcolor{conf3} 39.32 & \cellcolor{conf3} 1.02 & \cellcolor{conf3} 18.98 & \cellcolor{conf3} 40.68 \\
\midrule
\multirow{4}{*}{\textbf{Claude-4.6-Sonnet}}
& 0-shot & \cellcolor{ans0} 37.61 & \cellcolor{ind0} 24.89 & \cellcolor{conf0} 20.50 & \cellcolor{conf0} 4.39 & \cellcolor{conf0} 17.12 & \cellcolor{conf0} 58.00 \\
& 1-shot & \cellcolor{ans1} 34.46 & \cellcolor{ind1} 20.16 & \cellcolor{conf1} 17.91 & \cellcolor{conf1} 2.25 & \cellcolor{conf1} 16.55 & \cellcolor{conf1} 63.29 \\
& 2-shot & \cellcolor{ans2} 34.68 & \cellcolor{ind2} 20.05 & \cellcolor{conf2} 17.91 & \cellcolor{conf2} 2.14 & \cellcolor{conf2} 16.78 & \cellcolor{conf2} 63.18 \\
& 3-shot & \cellcolor{ans3} 36.04 & \cellcolor{ind3} 21.51 & \cellcolor{conf3} 19.14 & \cellcolor{conf3} 2.36 & \cellcolor{conf3} 16.89 & \cellcolor{conf3} 61.60 \\
\midrule
\multirow{4}{*}{\textbf{Qwen3.5-397B-A17B}}
& 0-shot & \cellcolor{ans0} 42.00 & \cellcolor{ind0} 24.89 & \cellcolor{conf0} 23.76 & \cellcolor{conf0} 1.13 & \cellcolor{conf0} 18.24 & \cellcolor{conf0} 56.87 \\
& 1-shot & \cellcolor{ans1} 44.85 & \cellcolor{ind1} 27.63 & \cellcolor{conf1} 24.12 & \cellcolor{conf1} 3.51 & \cellcolor{conf1} 20.73 & \cellcolor{conf1} 51.64 \\
& 2-shot & \cellcolor{ans2} 43.50 & \cellcolor{ind2} 27.80 & \cellcolor{conf2} 24.98 & \cellcolor{conf2} 2.83 & \cellcolor{conf2} 18.53 & \cellcolor{conf2} 53.67 \\
& 3-shot & \cellcolor{ans3} 42.72 & \cellcolor{ind3} 26.90 & \cellcolor{conf3} 24.19 & \cellcolor{conf3} 2.71 & \cellcolor{conf3} 18.53 & \cellcolor{conf3} 54.47 \\
\midrule
\multirow{4}{*}{\textbf{Qwen3VL-8B}}
& 0-shot & \cellcolor{ans0} 26.35 & \cellcolor{ind0} 4.39 & \cellcolor{conf0} 4.05 & \cellcolor{conf0} 0.34 & \cellcolor{conf0} 22.30 & \cellcolor{conf0} 73.31 \\
& 1-shot & \cellcolor{ans1} 25.23 & \cellcolor{ind1} 2.70 & \cellcolor{conf1} 2.59 & \cellcolor{conf1} 0.11 & \cellcolor{conf1} 22.64 & \cellcolor{conf1} 74.66 \\
& 2-shot & \cellcolor{ans2} 27.32 & \cellcolor{ind2} 3.67 & \cellcolor{conf2} 3.56 & \cellcolor{conf2} 0.11 & \cellcolor{conf2} 23.76 & \cellcolor{conf2} 72.57 \\
& 3-shot & \cellcolor{ans3} 27.50 & \cellcolor{ind3} 3.15 & \cellcolor{conf3} 3.15 & \cellcolor{conf3} 0.05 & \cellcolor{conf3} 24.35 & \cellcolor{conf3} 72.45 \\
\midrule
\multirow{4}{*}{\textbf{Qwen2VL-7B}}
& 0-shot & \cellcolor{ans0} 22.97 & \cellcolor{ind0} 0.45 & \cellcolor{conf0} 0.11 & \cellcolor{conf0} 0.34 & \cellcolor{conf0} 22.86 & \cellcolor{conf0} 76.69 \\
& 1-shot & \cellcolor{ans1} 30.41 & \cellcolor{ind1} 0.34 & \cellcolor{conf1} 0    & \cellcolor{conf1} 0.34 & \cellcolor{conf1} 30.41 & \cellcolor{conf1} 69.26 \\
& 2-shot & \cellcolor{ans2} 21.28 & \cellcolor{ind2} 0.90 & \cellcolor{conf2} 0    & \cellcolor{conf2} 0.90 & \cellcolor{conf2} 21.28 & \cellcolor{conf2} 77.82 \\
& 3-shot & \cellcolor{ans3} 26.35 & \cellcolor{ind3} 1.13 & \cellcolor{conf3} 0.68 & \cellcolor{conf3} 0.45 & \cellcolor{conf3} 25.68 & \cellcolor{conf3} 73.20 \\
\bottomrule
\end{tabular}%
}
\end{minipage}

\vspace{0.6cm}

\begin{minipage}[htbp]{0.40\textwidth}
    \centering
    \vspace{0pt}
    \includegraphics[width=\textwidth]{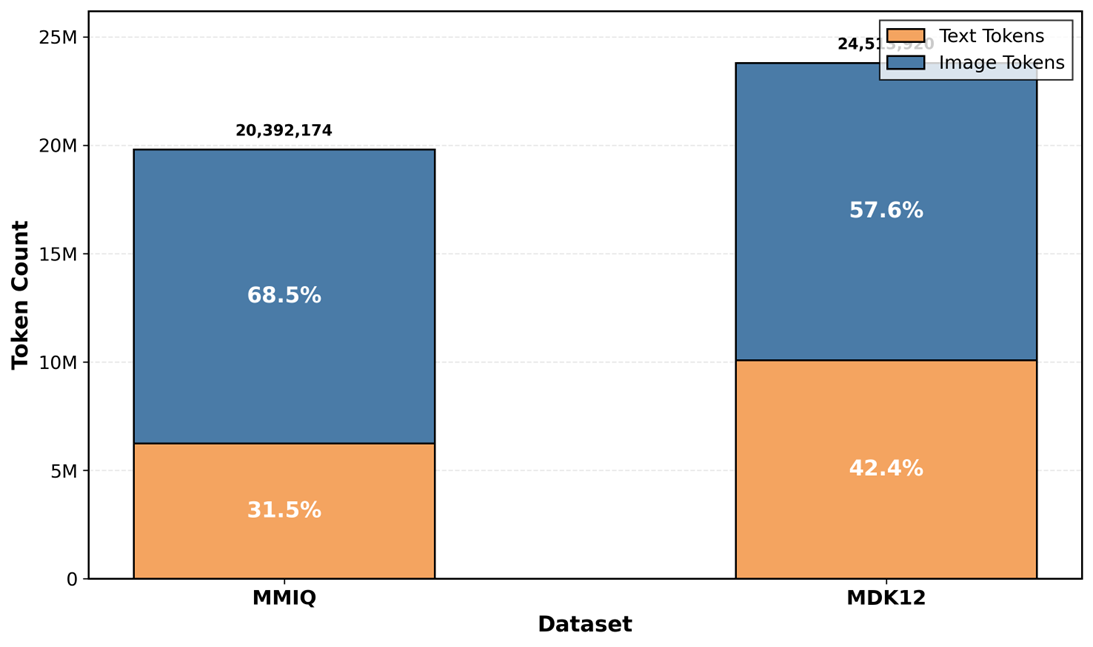}
    \captionof{figure}{Token count distribution between visual and textual tokens in multi-image ICL settings. The proportion of visual tokens is higher than that of text.}
    \label{fig:token_ratio}
\end{minipage}%
\hfill
\begin{minipage}[htbp]{0.58\textwidth}
    \centering
    \vspace{0pt}
    \includegraphics[width=\textwidth]{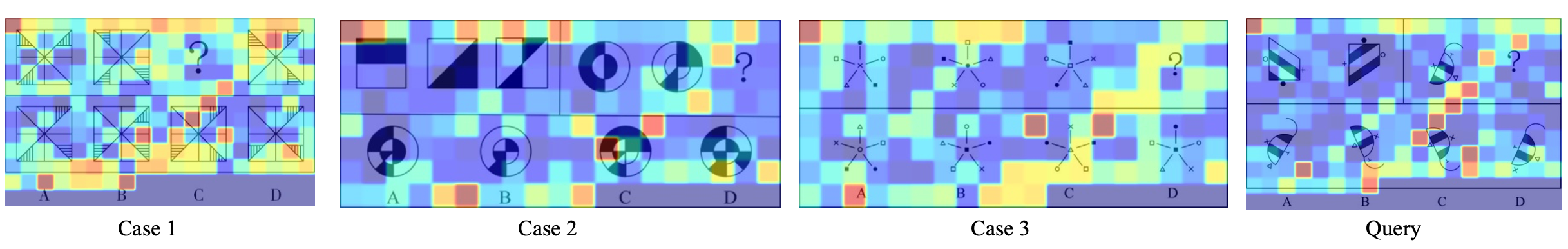}\\
    \includegraphics[width=\textwidth]{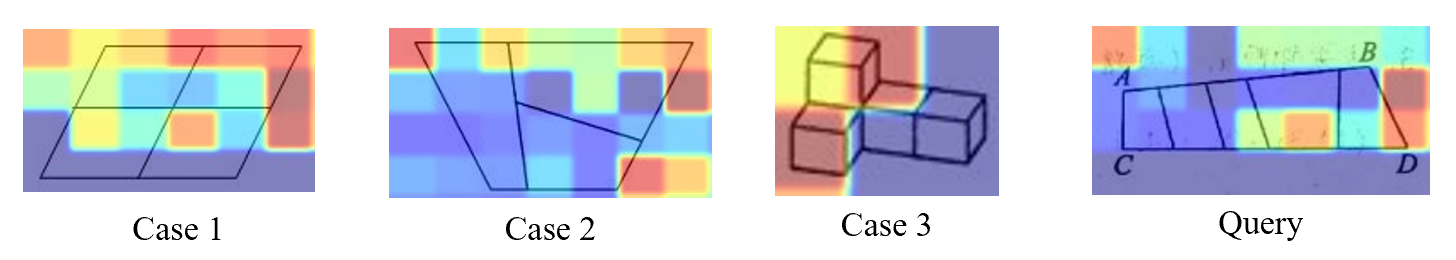}
    \captionof{figure}{Attention heatmap visualization for VLMs during ICL. The model's attention is disproportionately concentrated on the first image, while middle cases and the query receive insufficient attention.}
    \label{fig:attention}
\end{minipage}

\vspace{-1em}
\end{figure*}

The results reveal a clear divergence. On visual perception tasks (VQA), both closed-source and open-source VLMs exhibit consistent accuracy gains with more cases, confirming that the standard ICL paradigm works for shallow visual understanding. However, on logical reasoning (MMIQ), the benefits vanish. For open-source models like Qwen3-VL-8B, additional examples invert the expected ICL effect: accuracy monotonically declines from 32.30\% (0-shot) to 27.22\% (3-shot).  Even the advanced model, GPT-5.4, drops from 37.38\% to 35.66\%. 

\textbf{Why does multimodal ICL fail on reasoning‑intensive tasks?} To answer this, we directly evaluate the inductive abilities of VLMs. For MMIQ, which possesses a standard solution pattern (i.e., an underlying logical rule) for each question, we manually annotated 200 samples by labeling each question with its correct rule as the ground truth. To measure induction explicitly, we prompted the models to reference the given cases and generate thinking. Three high-performance models \cite{singh2025openai,pichai2025gemini3,anthropic2025claude} were used to vote and examine whether the VLM induced the correct rule, results are presented in Table~\ref{tab:induction}. Across all few-shot settings, the evaluated models exhibit a clear performance gap between answer accuracy and induction accuracy. For instance, under the 1-shot setting, only 2.70\% of Qwen3-VL-8B's outputs exhibit genuine inductive reasoning, while 25.23\% reach the right answer through flawed logic ($\times\rightarrow \checkmark$).  Even for the advanced closed-source model, Gemini 3.1-pro, approximately 16–19\% of answers remain "lucky guesses" across all settings. These results strongly indicate that existing VLMs substantially lack the ability to extract and apply logical rules from context.

\textbf{What prevents VLMs from performing effective induction?} We identify two critical obstacles at the representation level that impair the visual foundation necessary for cross‑case induction. First, we observe that during the decoding stage of the language model backbone, there exists a pronounced imbalance between visual and textual tokens under multi-image settings, as illustrated in Figure~\ref{fig:token_ratio}. In MDK12, visual tokens account for 57.6\% of the total, while this proportion reaches 68.5\% in MMIQ. This imbalance dilutes the textual evidence essential for logical induction, as the language model's computation is disproportionately consumed by redundant visual patches. Second, we extract attention weights based on the token position intervals of images in the sequence and visualize them as heatmaps. Specifically, for the effective attention tensor of each image $A \in \mathbb{R}^{H \times Q \times K}$ (where $H$, $Q$, $K$ denote the number of attention heads, query sequence length, and image token count, respectively), we first take the maximum along the query dimension, then average across the multi-head dimension to obtain a one-dimensional image token attention score:
\begin{equation}
    S = \frac{1}{H} \sum_{h=1}^{H} \max_{q \in Q} A_{h,q,:}
\end{equation}
After normalization, we reshape the one-dimensional scores into a two-dimensional spatial matrix according to the model's patch size. Results are shown in Figure~\ref{fig:attention}, which demonstrate that during ICL, VLM attention is heavily concentrated on the first in-context image, with subsequent cases and the query receiving markedly insufficient focus. This bias prevents the model from extracting additional reasoning gains from 2-shot and 3-shot settings.

In summary, we identify the hierarchical failure in multimodal ICL: at the cognitive level, the inductive correctness gap as the central bottleneck; at the representation level, visual token redundancy and attention imbalance that obstruct the induction. Eliminating these barriers is therefore a prerequisite for enabling genuinely inductive multimodal ICL, which motivates our framework design.

\section{Method}
\label{sec:method}




\begin{figure}[t]
    \centering
    \includegraphics[width=\textwidth]{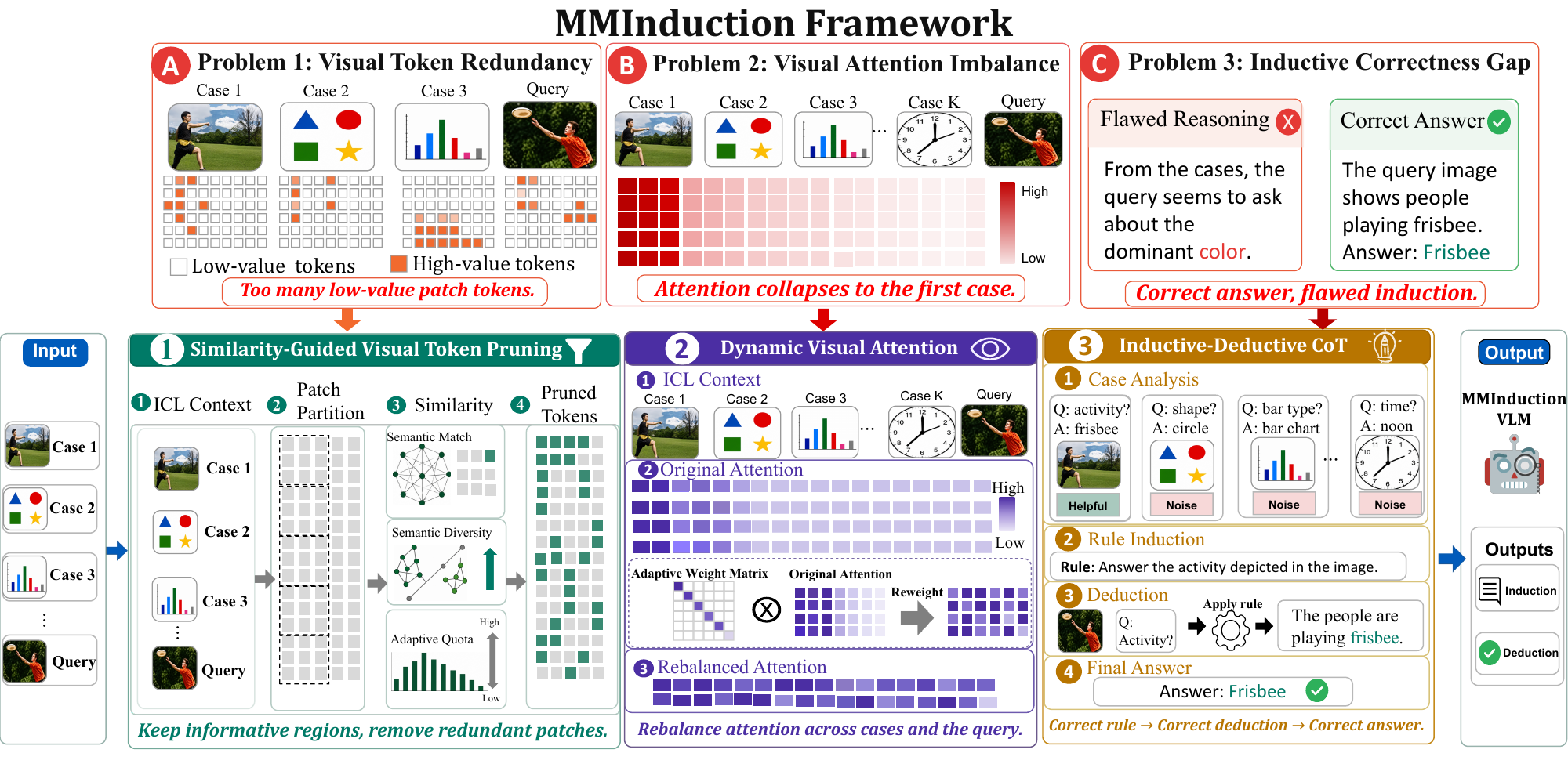}
    \caption{Overview of the proposed MMInduction framework. (1) Similarity-Guided Visual Token Pruning, which compresses redundant background patches while preserving key semantics; (2) Dynamic Visual Attention, which rebalances the model's attention distribution across all reference cases and the query; and (3) Inductive-Deductive CoT, which explicitly guides the model to first induce underlying rules from the helpful cases and subsequently deduce the final answer.}
    \label{fig:framework}
    \vspace{-1em}
\end{figure}

\subsection{Inductive-Deductive Chain-of-Thought}
\label{sec:cot}

ICL is fundamentally an induction-deduction process \cite{wies2023learnability,cheng2024inductive}. However, as revealed by our motivation study, the central bottleneck in multimodal ICL is inductive correctness gap, where VLMs fail to genuinely extract and apply rules from cases. To close this gap, we design a structured reasoning paradigm that explicitly guides the model through the induction-deduction process that humans employ.

\textbf{Reasoning Template.} We decompose the reasoning process into three sequential stages: (1) Case Analysis, where the model examines each demonstration individually, identifying its key elements and solution logic; (2) Induction, where the model synthesizes observations across cases to extract a generalizable rule; and (3) Deduction, where the extracted rule is systematically applied to the query to derive the final answer. It is worth noting that, during early experiments, we found that cases retrieved based on similarity are not always helpful for induction. Consequently, we incorporate a usefulness discrimination step. The model must judge whether each retrieved case is helpful or is a noise case that should be disregarded during induction. During training data construction, hard negative cases that are dissimilar to the query are randomly inserted into the context as noise, and the model must be able to recognize such distractors and make accurate judgments. Under the 3-shot and 5-shot settings, the numbers of noise cases are one and two, respectively.  The complete template is detailed in Appendix Table \ref{tab:prompt_induction}.

\textbf{Training Strategy.} To equip the model with this structured reasoning, we adopt a combined SFT + RLVR paradigm \cite{guo2025deepseek}. In the SFT stage, we distill high-quality reasoning  from Gemini 3.1-pro \cite{pichai2025gemini3}, as it demonstrates the best performance in preliminary experiments. Rather than generating the entire response end-to-end, we prompt the teacher to sequentially generate the case analysis, induction, and deduction steps. The synthesized reasoning data are then reviewed and voted upon by three expert models \cite{pichai2025gemini3,seed2026seed1,anthropic2025claude}, only samples securing majority agreement on logical validity are retained. 

In the RLVR stage, we design three fine-grained rule-based reward functions to guide the model toward correct reasoning while minimizing computational overhead:
\begin{equation}
\small
\begin{aligned}
R_{\text{task}}&=
\begin{cases}
1, & y_i=y_i^{gt} \\
0, & y_i\neq y_i^{gt} \\
-1, & \text{wrong format}
\end{cases}
\quad
R_{\text{helpful}}=
\begin{cases}
1, & h_j=\text{False} \\
0, & \text{otherwise}
\end{cases}
\quad
R_{\text{cite}}=
\begin{cases}
1, & \forall i,\ h_i=\text{True}\Rightarrow \text{cited} \ i \\
0, & \text{otherwise}
\end{cases}
\end{aligned}
\end{equation}
where $h_j$ represents whether the model deems the $j$-th case to be helpful, $R_\text{task}$ regulates the model's generation format and checks final result correctness, $R_\text{helpful}$ determines whether the model identifies noise cases, and $R_\text{cite}$ regulates the inductive-deductive process by requiring the model to fully cite samples previously identified as \textit{helpful} in its analysis. The final reward is a weighted sum:
\begin{equation}
    R = \alpha R_\text{task} + \beta R_\text{helpful} + \gamma R_\text{cite}
\end{equation}
where $\alpha = 1$, $\beta = 0.25$, $\gamma = 0.25$. The dominant weight on $R_{\text{task}}$ reflects that answer correctness remains the primary objective, while the auxiliary rewards provide dense intermediate supervision that guides the model toward faithful inductive-deductive reasoning. For policy optimization, we adopt DAPO \cite{yu2025dapo}. Our structured CoT produces substantially longer generations than typical QA tasks, which causes vanilla GRPO \cite{shao2024deepseekmath} to frequently exceed context length limits and terminate training prematurely. DAPO's built-in overlong sequence filtering mechanism effectively handles this issue, ensuring stable training. The policy is optimized with the following objective:
\begin{equation}
\begin{aligned}
&\mathcal{L}_{\mathrm{DAPO}}(\theta)
= {} \mathbb{E}_{(q,a)\sim \mathcal{D},\, \{o_i\}_{i=1}^{G}\sim \pi_{\theta_{\mathrm{old}}}(\cdot \mid q)}
\\ 
&\Biggl[
\frac{1}{\sum_{i=1}^{G} |o_i|}
\sum_{i=1}^{G}\sum_{t=1}^{|o_i|}
\min\!\left(
r_{i,t}(\theta)\hat{A}_{i,t},
\operatorname{clip}\!\left(r_{i,t}(\theta),\, 1-\epsilon_{\mathrm{low}},\, 1+\epsilon_{\mathrm{high}}\right)\hat{A}_{i,t}
\right) -\beta \mathbb{D}_{\mathrm{KL}}\left(\pi_\theta \,\|\, \pi_{\mathrm{ref}}\right)
\Biggr]
\end{aligned}
\end{equation}
where $r_{i,t}(\theta)
=
\frac{
\pi_{\theta}(o_{i,t} \mid q, o_{i,<t})
}{
\pi_{\theta_{\mathrm{old}}}(o_{i,t} \mid q, o_{i,<t})
}$, $\hat{A}_{i,t}
=
\frac{
R_i - \mathrm{mean}\left(\{R_i\}_{i=1}^{G}\right)
}{
\mathrm{std}\left(\{R_i\}_{i=1}^{G}\right)
}$, $\pi_{\theta}$ is the current policy, $\pi_{\text{old}}$ the sampling policy, and $\pi_{\text{ref}}$ a fixed reference model.

The SFT stage uses 15\% of the training data for cold-starting the structured reasoning format; the remaining 85\% is used for RLVR, where the model refines its inductive-deductive capability under the fine-grained composite reward. This design ensures that the model does not explicitly memorize reasoning patterns but instead internalizes the induction-deduction process as a generalizable cognitive routine.

\subsection{Similarity-Guided Visual Token Pruning}
\label{sec:pruning}

The Inductive-Deductive CoT relies on the model's ability to extract both textual and visual patterns across multiple cases. However, in multi-image settings, visual tokens dominate the input sequence (as shown in Figure \ref{fig:token_ratio}), diluting the textual evidence essential for induction. The goal of this module is therefore to compress visual inputs before they enter the language backbone, preserving semantic content while reducing the volume of tokens that compete with text for model attention.

A natural approach would be to compute token importance using attention scores, as done in several prior works \cite{li2026catp,tian2025identifying}. However, attention-based pruning entangles the compression with the model's current attention distribution, which is biased in multi-image settings. Thus, inspired by HoloV \cite{zou2025don}, we propose a similarity-guided visual token pruning strategy that compresses the visual input scale while preserving key semantic information.  Let an image be represented by \(N\) visual tokens. We first partition them uniformly into \(Z\) local spatial regions, each containing \(M\) tokens. For the \(z\)-th region with token representations \(\mathbf{E}^z \in \mathbb{R}^{M \times d}\), we compute the intra-region token similarity matrix:
\begin{equation}
    \mathbf{S}^z = (1 - \mathbf{I}_M) \odot (\mathbf{E}^z \mathbf{E}^{z^\top})
\end{equation}
where \(\mathbf{I}_M\) is the identity matrix that removes self-similarity. For the \(i\)-th token in region \(z\), its local semantic diversity is defined as the variance of its similarities to all other tokens in the same region:
\begin{equation}
    \nu_i^z = \frac{1}{M-1} \sum_j (\mathbf{S}_{i,j}^z - \mu_i^z)^2
\end{equation}
where \(\mu_i^z\) is the mean similarity of token \(i\) with other tokens in the region. A higher \(\nu_i^z\) indicates that the token carries distinctive information within its local context—it differs from its neighbors and is thus more likely to be semantically informative. We then allocate a retention budget across regions based on their aggregate importance. The weight for region \(z\) is:
\begin{equation}
    w_z = \frac{\left( \frac{1}{M} \sum_{t=1}^{M} \nu_t^z \right)^\tau}{\sum_{z'=1}^{Z} \left( \frac{1}{M} \sum_{t=1}^{M} \nu_t^{z'} \right)^\tau}
\end{equation}
where the sharpness parameter \(\tau\) controls how strongly the allocation favors high-diversity regions. Given a target of \(\hat{N}_v\) retained tokens per image, the initial quota for region \(z\) is \(q_z = \lfloor w_z \hat{N}_v \rfloor\). To guarantee \(\hat{N}_v\) tokens are preserved, we distribute the remaining \(R = \hat{N}_v - \sum_{z} q_z\) tokens according to the largest fractional remainders. The regions are ranked by \(w_z \hat{N}_v - q_z\), and the top-\(R\) regions each receive one additional token.  Within each region, we retain the \(q_z\) tokens with the highest diversity scores:
\begin{equation}
   \arg\max_{\Omega_z \subset \{1,\dots,M\}} \sum_{j \in \Omega_z} \nu_j^z, \quad \text{s.t.} \quad |\Omega_z| = q_z 
\end{equation}
In our implementation, we set \(Z = 16\) spatial regions and retain \(\hat{N}_v = 128\) tokens per image, reducing the original visual token count by approximately 80\% for VLM inputs. 
\subsection{Dynamic Visual Attention}
\label{sec:attention}

The Inductive-Deductive CoT requires the model to  compare and contrast multiple reference cases to extract a generalizable rule. However, as identified in our motivation study, pre-trained VLMs exhibit a severe visual attention imbalance in multi-image settings. To address this, inspired by DARA \cite{chen2025true}, we introduce a lightweight, context-conditioned dynamic visual attention mechanism that rebalances the attention distribution at the earliest stage of multimodal fusion.

Our design is guided by the observation that attention imbalance is image-specific rather than position-dependent. A simple scaling of attention on each image would be insufficient, as it ignores spatial variation within images. Conversely, naive per-token position biases, where each absolute sequence position receives a fixed learnable parameter, would fail to generalize across different inputs. We resolve this tension by exploiting a structural property of our pipeline. The preceding token pruning stage produces a fixed, canonical representation of \(P = 128\) visual tokens per image, regardless of its original resolution. This standardization transforms the visual input into a stable spatial canvas, making token-level modulation both meaningful and generalizable.

Concretely, for a multi-image context containing \(K\) images, we associate each image index \(i \in \{1, \dots, K\}\) with a learnable vector \(\mathbf{u}^{(i)} \in \mathbb{R}^{P}\) of per‑token scaling factors. Let the \(k\)-th visual token of image \(i\) reside at sequence position \(j\) (denoted \(j \in \mathcal{I}_{i,k}\)). The modulation factor for position \(j\) is then defined as $f_j = (\mathbf{u}^{(i)})_k$. We intervene at the first layer of the VLM's language backbone. The self-attention module first projects the input \(\mathbf{X}\) into queries, keys, and values, and computes the standard pre-softmax attention score matrix \(\mathbf{S} = \mathbf{Q}\mathbf{K}^\top\). We form a diagonal scaling matrix \(\mathbf{F} = \operatorname{diag}(f_1, f_2, \dots)\) and obtain the rebalanced scores by right‑multiplication: \(\mathbf{S}' = \mathbf{S} \mathbf{F}\). Because the scaling is applied before the softmax normalization, it directly shapes the final attention probability distribution rather than serving as a simple additive bias.

Textual tokens are entirely excluded from modulation (\(f_j = 1\)), ensuring that the mechanism purely redistributes attention across visual sources without distorting linguistic processing.  At initialization, all entries of all \(\mathbf{u}^{(i)}\) are set to 1, making the intervention a transformation that gradually learns to rebalance attention. During inference with a varying number of shots, we simply use the first \(K\) trained vectors, without any architectural adaptation.

\section{Experiments}
\label{sec:experiments}

\subsection{Experimental Settings}
\paragraph{Dataset.} To evaluate the proposed approach, we select eight datasets: VQAv2 \cite{antol2015vqa}, TextVQA \cite{singh2019towards}, MMIQ \cite{cai2025mm}, MME Reasoning \cite{yuan2025mme}, MDK12 \cite{zhou2026mdk12}, CSVQA \cite{jian2025csvqa}, MAMI \cite{fersini2022semeval}, and Harmeme \cite{pramanick2021detecting}. These datasets correspond to four cognitive tasks, namely visual perception, logical reasoning, STEM knowledge, and sarcasm recognition, respectively.  The training data consists of subsets with the same proportion of 1-shot, 3-shot and 5-shot setting. We report accuracy as the evaluation metric, which is determined by comparing model responses against ground truth labels using the high performance closed source model as judger. Details are provided in the Appendix.

\paragraph{Implementation Details.} All training and evaluation are conducted using PyTorch on 8$\times$NVIDIA A100 GPUs. All the evaluations were repeated three times under different random seeds (42, 1895, 2026), and the average values are reported. Feature embeddings for the retrieval stage were generated using Qwen3VL-Embedding-8B  \cite{li2026qwen3}. The impact of different embedding models and retrieval strategies is compared in the Appendix Table. The closed source models used in evaluation are accessed via API calls on a CPU server. Training was conducted under the MS-Swift framework \cite{zhao2025swift}. The prompt templates used for training and evaluation, the hyperparameters in training, and case studies of both successful and failed instances are provided in the Appendix. The complete code, data, and partial logs are available in the supplementary material.

\subsection{Main Result}
\begin{table}[t]
\centering
\caption{Performance comparison across selected benchmarks. We evaluate methods under 1-shot and 3-shot settings. The $\Delta$  reports the relative improvement over the zero-shot baseline. \textbf{Bold} and \underline{underlined} values indicate the best and second-best results among all methods.}
\label{tab:main_result}
\begin{adjustbox}{max width=\textwidth}
\begin{tabular}{lcccccccccccc}
\toprule
\multirow{2}{*}{\diagbox[width=9em, height=2.5\line]{\textbf{Method}}{\textbf{Benchmark}}}
  & \multicolumn{3}{c}{\textbf{VQAv2}}
  & \multicolumn{3}{c}{\textbf{MMIQ}}
  & \multicolumn{3}{c}{\textbf{MDK12}}
  & \multicolumn{3}{c}{\textbf{MAMI}} \\
\cmidrule(lr){2-4} \cmidrule(lr){5-7} \cmidrule(lr){8-10} \cmidrule(lr){11-13}
  & 1-shot & 3-shot & $\Delta$
  & 1-shot & 3-shot & $\Delta$
  & 1-shot & 3-shot & $\Delta$
  & 1-shot & 3-shot & $\Delta$ \\
\midrule
\multicolumn{13}{c}{\textbf{Qwen2-VL-7B-Instruct}} \\
\midrule
Zero-shot  & \multicolumn{2}{c}{66.92} & --  & \multicolumn{2}{c}{26.56} & --  & \multicolumn{2}{c}{28.96} & --  & \multicolumn{2}{c}{57.50} & -- \\
\cdashline{1-13}
Vanilla & 69.54 & 71.64 & \textcolor{green!60!black}{+4.72\%} & 25.50 & 27.45 & \textcolor{green!60!black}{+0.89\%} & \underline{33.99} & 34.33 & \textcolor{green!60!black}{+5.37\%} & 65.00 & 75.00 & \textcolor{green!60!black}{+17.50\%} \\
SoFA & 70.35 & 72.76 & \textcolor{green!60!black}{+5.84\%} & 26.74 & 28.06 & \textcolor{green!60!black}{+1.50\%} & \textbf{34.14} & 35.21 & \textcolor{green!60!black}{+6.25\%} & 66.41 & 75.58 & \textcolor{green!60!black}{+18.08\%} \\
LoRA & 75.21 & 76.07 & \textcolor{green!60!black}{+9.15\%} & \underline{26.87} & 27.56 & \textcolor{green!60!black}{+1.00\%} & 32.34 & 34.27 & \textcolor{green!60!black}{+5.31\%} & 75.00 & 78.33 & \textcolor{green!60!black}{+20.83\%} \\
DARA & \underline{76.66} & \underline{77.89} & \textcolor{green!60!black}{\underline{+10.97\%}} & 26.45 & \underline{28.77} & \textcolor{green!60!black}{\underline{+2.21\%}} & 33.90 & \underline{36.18} & \textcolor{green!60!black}{\underline{+7.22\%}} & \underline{76.67} & \textbf{79.57} & \textcolor{green!60!black}{\textbf{+22.07\%}} \\
\rowcolor{gray!20} MMInduction & \textbf{78.45} & \textbf{80.22} & \textcolor{green!60!black}{\textbf{+13.30\%}} & \textbf{27.99} & \textbf{30.01} & \textcolor{green!60!black}{\textbf{+3.45\%}} & 33.39 & \textbf{38.79} & \textcolor{green!60!black}{\textbf{+9.83\%}} & \textbf{78.54} & \underline{79.44} & \textcolor{green!60!black}{\underline{+21.94\%}} \\
\midrule
\multicolumn{13}{c}{\textbf{Qwen2.5-VL-7B-Instruct}} \\
\midrule
Zero-shot  & \multicolumn{2}{c}{73.65} & --  & \multicolumn{2}{c}{25.93} & --  & \multicolumn{2}{c}{26.39} & --  & \multicolumn{2}{c}{72.50} & -- \\
\cdashline{1-13}
Vanilla & 74.64 & 78.35 & \textcolor{green!60!black}{+4.70\%} & 20.59 & 24.38 & \textcolor{red!70!black}{-1.55\%} & 27.72 & 28.46 & \textcolor{green!60!black}{+2.07\%} & 76.67 & 79.17 & \textcolor{green!60!black}{+6.67\%} \\
SoFA & 75.85 & 79.19 & \textcolor{green!60!black}{+5.54\%} & 20.24 & 25.54 & \textcolor{red!70!black}{-0.39\%} & 28.64 & 28.28 & \textcolor{green!60!black}{+2.25\%} & 77.56 & 80.31 & \textcolor{green!60!black}{+7.81\%} \\
LoRA & 85.40 & 85.47 & \textcolor{green!60!black}{+11.82\%} & 24.03 & 26.27 & \textcolor{green!60!black}{+0.34\%} & 34.89 & 36.10 & \textcolor{green!60!black}{+9.71\%} & 82.50 & 85.00 & \textcolor{green!60!black}{+12.50\%} \\
DARA & \underline{86.92} & \underline{87.35} & \textcolor{green!60!black}{\underline{+13.70\%}} & \underline{25.78} & \underline{27.46} & \textcolor{green!60!black}{\underline{+1.53\%}} & \textbf{36.55} & \underline{37.18} & \textcolor{green!60!black}{\underline{+10.79\%}} & \underline{84.04} & \underline{86.12} & \textcolor{green!60!black}{\underline{+13.62\%}} \\
\rowcolor{gray!20} MMInduction & \textbf{88.81} & \textbf{89.72} & \textcolor{green!60!black}{\textbf{+16.07\%}} & \textbf{28.15} & \textbf{29.85} & \textcolor{green!60!black}{\textbf{+3.92\%}} & \underline{36.13} & \textbf{40.02} & \textcolor{green!60!black}{\textbf{+13.63\%}} & \textbf{86.92} & \textbf{88.18} & \textcolor{green!60!black}{\textbf{+15.68\%}} \\
\midrule
\multicolumn{13}{c}{\textbf{Qwen3-VL-8B-Instruct}} \\
\midrule
Zero-shot  & \multicolumn{2}{c}{83.62} & --  & \multicolumn{2}{c}{32.30} & --  & \multicolumn{2}{c}{46.59} & --  & \multicolumn{2}{c}{55.00} & -- \\
\cdashline{1-13}
Vanilla & 84.05 & 84.47 & \textcolor{green!60!black}{+0.85\%} & 28.70 & 27.22 & \textcolor{red!70!black}{-3.60\%} & 48.47 & 48.98 & \textcolor{green!60!black}{+2.39\%} & 83.33 & 85.83 & \textcolor{green!60!black}{+30.83\%} \\
SoFA & 84.87 & 85.82 & \textcolor{green!60!black}{+2.20\%} & 29.61 & 28.66 & \textcolor{red!70!black}{-2.69\%} & 48.22 & 49.87 & \textcolor{green!60!black}{+3.28\%} & 84.54 & \underline{86.77} & \textcolor{green!60!black}{\underline{+31.77\%}} \\
LoRA & 83.76 & 84.90 & \textcolor{green!60!black}{+1.28\%} & \underline{34.21} & 35.63 & \textcolor{green!60!black}{+3.33\%} & 50.09 & 52.97 & \textcolor{green!60!black}{+6.38\%} & 83.33 & 83.72 & \textcolor{green!60!black}{+28.72\%} \\
DARA & \underline{85.40} & \underline{86.15} & \textcolor{green!60!black}{\underline{+2.53\%}} & 33.69 & \underline{37.39} & \textcolor{green!60!black}{\underline{+5.09\%}} & \underline{51.66} & \underline{54.79} & \textcolor{green!60!black}{\underline{+8.20\%}} & \underline{85.18} & 84.80 & \textcolor{green!60!black}{+30.18\%} \\
\rowcolor{gray!20} MMInduction & \textbf{86.88} & \textbf{89.46} & \textcolor{green!60!black}{\textbf{+5.84\%}} & \textbf{35.73} & \textbf{38.46} & \textcolor{green!60!black}{\textbf{+6.16\%}} & \textbf{54.90} & \textbf{55.51} & \textcolor{green!60!black}{\textbf{+8.92\%}} & \textbf{86.74} & \textbf{88.05} & \textcolor{green!60!black}{\textbf{+33.05\%}} \\
\bottomrule
\end{tabular}
\end{adjustbox}
\end{table}

We select Qwen2-VL-7B \cite{wang2024qwen2}, Qwen2.5-VL-7B \cite{bai2025qwen2_5}, and Qwen3VL-8B \cite{bai2025qwen3} as the base models for training. We compare vanilla VLMs under 1-shot and 3-shot settings with multimodal ICL baselines including SoFA \cite{tian2025identifying}, LoRA \cite{hu2022lora}, and DARA \cite{chen2025true}. Specifically, SoFA is a training-free visual attention adjustment. LoRA denotes parameter-efficient SFT. DARA is a visual attention balancing SFT strategy designed for multimodal ICL. LoRA and DARA are fine-tuned using 15\% data with Inductive CoT and 85\% data with QA pair, ensuring that the amount of data used is consistent with MMInduction.

The results are illustrated in Table \ref{tab:main_result}. Vanilla VLMs exhibit a performance inversion on logical reasoning benchmarks like MMIQ. SoFA yields unstable, marginal gains, while LoRA and DARA bring only limited improvements, indicating that attention modulation alone cannot overcome the core inductive deficiency. In contrast, MMInduction achieves robust and consistent gains across all datasets. The gains are not attributable to increased data or computation alone, as the same training data and similar parameter budgets applied to LoRA/DARA fail to close the inductive gap.  This demonstrates that genuine multimodal ICL can be activated only when models are both instructed to induce rules and equipped with a balanced, concise visual foundation. The complete results on eight datasets are presented in Appendix Table \ref{tab:main_result_appendix}.

\subsection{Explainability Analysis}

The explainability results in Table 4 confirm that MMInduction remedies the central deficiency diagnosed in our motivation study, i.e.  the inductive correctness gap.  After MMInduction training, both answer accuracy and induction accuracy rise as the number of reference cases increases from one to three shots, with induction accuracy reaching 20.78\%. This growth demonstrates that the framework genuinely equips the model with cross‑case rule extraction, not merely answer memorization. However, extending to four and five shots causes a decline in the success rate. The model begins to over‑analyze cases, producing excessively long responses that exceed the context limit, triggering reasoning failures. This reveals that while our inductive CoT imparts the capability for rule induction, the base model’s raw reasoning stability remains a bottleneck. Thus,  it is necessary to select cases properly to balance inductive depth with generation stability.

\begin{figure}[ht]
\centering
\begin{minipage}[t]{0.48\textwidth}
\centering
\captionof{table}{Induction Analysis on Qwen3VL-8B.}
\label{tab:hyper}
\resizebox{\linewidth}{!}{%
\begin{tabular}{lccc}
\toprule
Setting & Answer Acc. & Induction Acc. & SR \\
\midrule
Vanilla@0-shot     & 26.35 & 4.39  & 94.59\% \\
Ours@0-shot & 24.66 & 3.42  & 100\%   \\
Ours@1-shot             & 35.73 & 16.25  & 100\%   \\
Ours@2-shot             & 37.58 & 18.93  & 100\%   \\
Ours@3-shot             & 38.46 & 20.78 & 91.75\%   \\
Ours@4-shot             & \textbf{39.12} & 21.54 & 83.55\% \\
Ours@5-shot             & 38.27 & \textbf{22.16}  & 69.48\% \\
\bottomrule
\end{tabular}
}
\end{minipage}%
\hfill
\begin{minipage}[t]{0.48\textwidth}
\centering
\captionof{table}{Ablation Study on Qwen3VL-8B.}
\label{tab:ablation}
\resizebox{\linewidth}{!}{%
\begin{tabular}{lccc}
\toprule
Setting \textbackslash{} Benchmark & VQAv2 & MMIQ & MDK12 \\
\midrule
MMInduction@3-shot                             & \textbf{89.46} & \textbf{37.81} & \textbf{55.51} \\
\midrule
w/o Token Pruning                              & 87.12 & 36.54 & 53.27 \\
w/o Dynamic Attention                          & 86.73 & 35.08 & 51.64 \\
w/o Inductive CoT                              & 84.31 & 33.47 & 48.82 \\
w/o $R_{\text{helpful}}\ \&\ R_{\text{cite}}$  & 88.93 & 37.69 & 54.88 \\
Inductive CoT Only                             & 88.65 & 35.33 & 53.13 \\
\bottomrule
\end{tabular}}
\end{minipage}
\end{figure}

\subsection{Ablation Study}

The ablation experiments in Table \ref{tab:ablation} isolate the contribution of each component and provide direct causal evidence for the hierarchical bottleneck analysis presented in Section \ref{sec:motivation}. Removing Similarity‑Guided Visual Token Pruning causes a clear performance drop, confirming that the overwhelming visual token redundancy indeed dilutes textual evidence and hampers induction. Ablating Dynamic Visual Attention leads to an even sharper decline, verifying that the first‑image attention bias is a severe barrier to cross‑case comparison. The most drastic drop occurs when we eliminate the Inductive CoT and rely solely on RLVR with a task reward. Removing the auxiliary helpful‑identification and citation rewards moderately reduces performance, highlighting that fine‑grained supervision on noise filtering and explicit case citation enriches the training signal beyond mere answer correctness. Finally, retaining only the Inductive CoT without visual pruning or attention rebalancing yields a gain over the vanilla model but falls significantly short of the full framework, demonstrating that structured reasoning alone cannot compensate for a corrupted visual foundation.

\section{Conclusion}
\label{sec:conclusion}

In this work, we identify three critical bottlenecks in multimodal in-context learning for VLMs: visual attention imbalance, visual token redundancy, and insufficient inductive correctness. To address these challenges, we propose MMInduction, a comprehensive framework that enhances multimodal ICL through inductive-deductive reasoning. Our method integrates similarity-guided visual token pruning, dynamic visual attention, and an inductive-deductive chain-of-thought paradigm, trained via a combined SFT and RLVR pipeline with fine-grained reward functions. Extensive experiments on eight datasets across four task categories demonstrate the effectiveness of our approach.

\bibliographystyle{plain}
\bibliography{myref}

\appendix

\section{Supplementary Experiment}
\label{app:sup_exp}

\begin{sidewaystable}[htbp]
\centering
\caption{Performance comparison across different benchmarks and settings. For each model and dataset, we evaluate methods under 1-shot and 3-shot in-context learning settings. The $\Delta$ column reports the relative improvement of the better-performing shot setting over the zero-shot baseline, computed as $\Delta = (\max(\text{1-shot}, \text{3-shot}) - \text{Zero-shot}) / \text{Zero-shot} \times 100\%$. \textbf{Bold} and \underline{underlined} values indicate the best and second-best results among all methods for each model--dataset--setting combination, respectively.}
\label{tab:main_result_appendix}
\begin{adjustbox}{max width=\textheight}
\begin{tabular}{lcccccccccccccccccccccccc}
\toprule
\multirow{3}{*}{\textbf{}}
  & \multicolumn{6}{c}{\textbf{Visual Perception}}
  & \multicolumn{6}{c}{\textbf{Logical Reasoning}}
  & \multicolumn{6}{c}{\textbf{STEM Reasoning}}
  & \multicolumn{6}{c}{\textbf{Human Preference}} \\
\cmidrule(lr){2-7} \cmidrule(lr){8-13} \cmidrule(lr){14-19} \cmidrule(lr){20-25}
  & \multicolumn{3}{c}{\textbf{VQAv2}}  & \multicolumn{3}{c}{\textbf{TextVQA}}  & \multicolumn{3}{c}{\textbf{MMIQ}}  & \multicolumn{3}{c}{\textbf{MME-R}}  & \multicolumn{3}{c}{\textbf{MDK12}}  & \multicolumn{3}{c}{\textbf{CSVQA}}  & \multicolumn{3}{c}{\textbf{Harmeme}}  & \multicolumn{3}{c}{\textbf{MAMI}} \\
\cmidrule(lr){2-4} \cmidrule(lr){5-7} \cmidrule(lr){8-10} \cmidrule(lr){11-13} \cmidrule(lr){14-16} \cmidrule(lr){17-19} \cmidrule(lr){20-22} \cmidrule(lr){23-25}
  & 1-shot & 3-shot & $\Delta$  & 1-shot & 3-shot & $\Delta$  & 1-shot & 3-shot & $\Delta$  & 1-shot & 3-shot & $\Delta$  & 1-shot & 3-shot & $\Delta$  & 1-shot & 3-shot & $\Delta$  & 1-shot & 3-shot & $\Delta$  & 1-shot & 3-shot & $\Delta$ \\
\midrule
\multicolumn{25}{c}{\textbf{Qwen2-VL-7B-Instruct}} \\
\midrule
Zero-shot  & \multicolumn{2}{c}{66.92} & --  & \multicolumn{2}{c}{47.65} & --  & \multicolumn{2}{c}{26.56} & --  & \multicolumn{2}{c}{23.60} & --  & \multicolumn{2}{c}{28.96} & --  & \multicolumn{2}{c}{30.64} & --  & \multicolumn{2}{c}{34.89} & --  & \multicolumn{2}{c}{57.50} & -- \\
\cdashline{1-25}
Vanilla & 69.54 & 71.64 & \textcolor{green!60!black}{+7.05\%} & 52.35 & 55.62 & \textcolor{green!60!black}{+16.73\%} & 25.50 & 27.45 & \textcolor{green!60!black}{+3.35\%} & 19.57 & 22.73 & \textcolor{red!70!black}{-3.69\%} & \underline{33.99} & 34.33 & \textcolor{green!60!black}{+18.54\%} & 34.79 & 31.15 & \textcolor{green!60!black}{+13.54\%} & 59.81 & 59.19 & \textcolor{green!60!black}{+71.42\%} & 65.00 & 75.00 & \textcolor{green!60!black}{+30.43\%} \\
SoFA & 70.35 & 72.76 & \textcolor{green!60!black}{+8.73\%} & 52.77 & 55.41 & \textcolor{green!60!black}{+16.29\%} & 26.74 & 28.06 & \textcolor{green!60!black}{+5.65\%} & 20.52 & 24.21 & \textcolor{green!60!black}{+2.58\%} & \textbf{34.14} & 35.21 & \textcolor{green!60!black}{+21.58\%} & 34.25 & 32.47 & \textcolor{green!60!black}{+11.78\%} & 60.57 & 60.24 & \textcolor{green!60!black}{+73.60\%} & 66.41 & 75.58 & \textcolor{green!60!black}{+31.44\%} \\
LoRA & 75.21 & 76.07 & \textcolor{green!60!black}{+13.67\%} & 59.82 & 60.94 & \textcolor{green!60!black}{+27.89\%} & \underline{26.87} & 27.56 & \textcolor{green!60!black}{+3.77\%} & 19.62 & 24.67 & \textcolor{green!60!black}{+4.53\%} & 32.34 & 34.27 & \textcolor{green!60!black}{+18.34\%} & 35.74 & 35.53 & \textcolor{green!60!black}{+16.64\%} & 73.05 & \underline{76.01} & \textcolor{green!60!black}{\underline{+117.86\%}} & 75.00 & 78.33 & \textcolor{green!60!black}{+36.23\%} \\
DARA & \underline{76.66} & \underline{77.89} & \textcolor{green!60!black}{\underline{+16.39\%}} & \underline{60.97} & \underline{62.58} & \textcolor{green!60!black}{\underline{+31.33\%}} & 26.45 & \underline{28.77} & \textcolor{green!60!black}{\underline{+8.32\%}} & \underline{21.50} & \underline{25.72} & \textcolor{green!60!black}{\underline{+8.98\%}} & 33.90 & \underline{36.18} & \textcolor{green!60!black}{\underline{+24.93\%}} & \underline{37.08} & \underline{37.29} & \textcolor{green!60!black}{\underline{+21.70\%}} & \underline{74.17} & 75.46 & \textcolor{green!60!black}{+116.28\%} & \underline{76.67} & \textbf{79.57} & \textcolor{green!60!black}{\textbf{+38.38\%}} \\
\rowcolor{gray!20} MMInduction & \textbf{78.45} & \textbf{80.22} & \textcolor{green!60!black}{\textbf{+19.87\%}} & \textbf{62.38} & \textbf{64.75} & \textcolor{green!60!black}{\textbf{+35.89\%}} & \textbf{27.99} & \textbf{30.01} & \textcolor{green!60!black}{\textbf{+12.99\%}} & \textbf{23.83} & \textbf{27.81} & \textcolor{green!60!black}{\textbf{+17.84\%}} & 33.39 & \textbf{38.79} & \textcolor{green!60!black}{\textbf{+33.94\%}} & \textbf{38.62} & \textbf{38.94} & \textcolor{green!60!black}{\textbf{+27.09\%}} & \textbf{77.81} & \textbf{78.16} & \textcolor{green!60!black}{\textbf{+124.02\%}} & \textbf{78.54} & \underline{79.44} & \textcolor{green!60!black}{\underline{+38.16\%}} \\
\midrule
\multicolumn{25}{c}{\textbf{Qwen2.5-VL-7B-Instruct}} \\
\midrule
Zero-shot  & \multicolumn{2}{c}{73.65} & --  & \multicolumn{2}{c}{73.42} & --  & \multicolumn{2}{c}{25.93} & --  & \multicolumn{2}{c}{23.74} & --  & \multicolumn{2}{c}{26.39} & --  & \multicolumn{2}{c}{29.57} & --  & \multicolumn{2}{c}{35.51} & --  & \multicolumn{2}{c}{72.50} & -- \\
\cdashline{1-25}
Vanilla & 74.64 & 78.35 & \textcolor{green!60!black}{+6.38\%} & 71.57 & \underline{74.74} & \textcolor{green!60!black}{\underline{+1.80\%}} & 20.59 & 24.38 & \textcolor{red!70!black}{-5.98\%} & 24.46 & 26.20 & \textcolor{green!60!black}{+10.36\%} & 27.72 & 28.46 & \textcolor{green!60!black}{+7.84\%} & 32.45 & 33.53 & \textcolor{green!60!black}{+13.39\%} & 78.97 & 83.18 & \textcolor{green!60!black}{+134.24\%} & 76.67 & 79.17 & \textcolor{green!60!black}{+9.20\%} \\
SoFA & 75.85 & 79.19 & \textcolor{green!60!black}{+7.52\%} & 72.99 & \textbf{75.25} & \textcolor{green!60!black}{\textbf{+2.49\%}} & 20.24 & 25.54 & \textcolor{red!70!black}{-1.50\%} & 25.23 & 27.59 & \textcolor{green!60!black}{+16.22\%} & 28.64 & 28.28 & \textcolor{green!60!black}{+8.53\%} & \underline{33.50} & 35.01 & \textcolor{green!60!black}{+18.40\%} & 79.60 & 84.45 & \textcolor{green!60!black}{+137.82\%} & 77.56 & 80.31 & \textcolor{green!60!black}{+10.77\%} \\
LoRA & 85.40 & 85.47 & \textcolor{green!60!black}{+16.05\%} & 71.78 & 74.74 & \textcolor{green!60!black}{+1.80\%} & 24.03 & 26.27 & \textcolor{green!60!black}{+1.31\%} & 27.62 & 29.40 & \textcolor{green!60!black}{+23.84\%} & 34.89 & 36.10 & \textcolor{green!60!black}{+36.79\%} & 32.64 & 34.39 & \textcolor{green!60!black}{+16.30\%} & 82.24 & 86.45 & \textcolor{green!60!black}{+143.45\%} & 82.50 & 85.00 & \textcolor{green!60!black}{+17.24\%} \\
DARA & \underline{86.92} & \underline{87.35} & \textcolor{green!60!black}{\underline{+18.60\%}} & \underline{73.02} & 74.13 & \textcolor{green!60!black}{+0.97\%} & \underline{25.78} & \underline{27.46} & \textcolor{green!60!black}{\underline{+5.90\%}} & \underline{29.56} & \underline{30.82} & \textcolor{green!60!black}{\underline{+29.82\%}} & \textbf{36.55} & \underline{37.18} & \textcolor{green!60!black}{\underline{+40.89\%}} & 32.19 & \underline{36.20} & \textcolor{green!60!black}{\underline{+22.42\%}} & \underline{83.61} & \underline{88.40} & \textcolor{green!60!black}{\underline{+148.94\%}} & \underline{84.04} & \underline{86.12} & \textcolor{green!60!black}{\underline{+18.79\%}} \\
\rowcolor{gray!20} MMInduction & \textbf{88.81} & \textbf{89.72} & \textcolor{green!60!black}{\textbf{+21.82\%}} & \textbf{73.85} & 74.64 & \textcolor{green!60!black}{+1.66\%} & \textbf{28.15} & \textbf{29.85} & \textcolor{green!60!black}{\textbf{+15.12\%}} & \textbf{30.09} & \textbf{34.21} & \textcolor{green!60!black}{\textbf{+44.10\%}} & \underline{36.13} & \textbf{40.02} & \textcolor{green!60!black}{\textbf{+51.65\%}} & \textbf{35.19} & \textbf{38.55} & \textcolor{green!60!black}{\textbf{+30.37\%}} & \textbf{86.02} & \textbf{89.09} & \textcolor{green!60!black}{\textbf{+150.89\%}} & \textbf{86.92} & \textbf{88.18} & \textcolor{green!60!black}{\textbf{+21.63\%}} \\
\midrule
\multicolumn{25}{c}{\textbf{Qwen3-VL-8B-Instruct}} \\
\midrule
Zero-shot  & \multicolumn{2}{c}{83.62} & --  & \multicolumn{2}{c}{70.76} & --  & \multicolumn{2}{c}{32.30} & --  & \multicolumn{2}{c}{22.28} & --  & \multicolumn{2}{c}{46.59} & --  & \multicolumn{2}{c}{35.31} & --  & \multicolumn{2}{c}{38.55} & --  & \multicolumn{2}{c}{55.00} & -- \\
\cdashline{1-25}
Vanilla & 84.05 & 84.47 & \textcolor{green!60!black}{+1.02\%} & 71.57 & 73.52 & \textcolor{green!60!black}{+3.90\%} & 28.70 & 27.22 & \textcolor{red!70!black}{-11.15\%} & 26.92 & 28.87 & \textcolor{green!60!black}{+29.58\%} & 48.47 & 48.98 & \textcolor{green!60!black}{+5.13\%} & 35.41 & 38.43 & \textcolor{green!60!black}{+8.84\%} & 82.09 & 83.96 & \textcolor{green!60!black}{+117.80\%} & 83.33 & 85.83 & \textcolor{green!60!black}{+56.05\%} \\
SoFA & 84.87 & 85.82 & \textcolor{green!60!black}{+2.63\%} & 72.71 & 73.06 & \textcolor{green!60!black}{+3.25\%} & 29.61 & 28.66 & \textcolor{red!70!black}{-8.33\%} & 27.50 & 30.14 & \textcolor{green!60!black}{+35.28\%} & 48.22 & 49.87 & \textcolor{green!60!black}{+7.04\%} & 36.57 & 39.15 & \textcolor{green!60!black}{+10.88\%} & 83.54 & 84.64 & \textcolor{green!60!black}{+119.56\%} & 84.54 & \underline{86.77} & \textcolor{green!60!black}{\underline{+57.76\%}} \\
LoRA & 83.76 & 84.90 & \textcolor{green!60!black}{+1.53\%} & 73.42 & 75.88 & \textcolor{green!60!black}{+7.24\%} & \underline{34.21} & 35.63 & \textcolor{green!60!black}{+10.31\%} & 27.13 & 29.58 & \textcolor{green!60!black}{+32.76\%} & 50.09 & 52.97 & \textcolor{green!60!black}{+13.69\%} & 37.22 & \underline{39.61} & \textcolor{green!60!black}{\underline{+12.18\%}} & 83.49 & 89.64 & \textcolor{green!60!black}{+132.53\%} & 83.33 & 83.72 & \textcolor{green!60!black}{+52.22\%} \\
DARA & \underline{85.40} & \underline{86.15} & \textcolor{green!60!black}{\underline{+3.03\%}} & \underline{75.30} & \underline{77.29} & \textcolor{green!60!black}{\underline{+9.23\%}} & 33.69 & \underline{37.39} & \textcolor{green!60!black}{\underline{+15.76\%}} & \textbf{29.08} & \underline{30.72} & \textcolor{green!60!black}{\underline{+37.88\%}} & \underline{51.66} & \underline{54.79} & \textcolor{green!60!black}{\underline{+17.60\%}} & \underline{38.58} & 39.23 & \textcolor{green!60!black}{+11.10\%} & \underline{85.20} & \textbf{90.93} & \textcolor{green!60!black}{\textbf{+135.88\%}} & \underline{85.18} & 84.80 & \textcolor{green!60!black}{+54.87\%} \\
\rowcolor{gray!20} MMInduction & \textbf{86.88} & \textbf{89.46} & \textcolor{green!60!black}{\textbf{+6.98\%}} & \textbf{76.16} & \textbf{79.77} & \textcolor{green!60!black}{\textbf{+12.73\%}} & \textbf{35.73} & \textbf{38.46} & \textcolor{green!60!black}{\textbf{+19.07\%}} & \underline{28.55} & \textbf{33.25} & \textcolor{green!60!black}{\textbf{+49.24\%}} & \textbf{54.90} & \textbf{55.51} & \textcolor{green!60!black}{\textbf{+19.15\%}} & \textbf{41.14} & \textbf{43.76} & \textcolor{green!60!black}{\textbf{+23.93\%}} & \textbf{86.17} & \underline{90.49} & \textcolor{green!60!black}{\underline{+134.73\%}} & \textbf{86.74} & \textbf{88.05} & \textcolor{green!60!black}{\textbf{+60.09\%}} \\
\bottomrule
\end{tabular}
\end{adjustbox}
\end{sidewaystable}

To comprehensively evaluate the robustness and generalizability of our proposed framework, Table \ref{tab:main_result_appendix} presents the complete experimental results across eight diverse datasets spanning four distinct cognitive domains: visual perception, logical reasoning, STEM reasoning, and human preference. Consistent with our preliminary observations, vanilla VLMs exhibit severe performance degradation in multi-shot settings for reasoning-intensive tasks (e.g., MMIQ), underscoring their inability to perform genuine cross-case induction. While baseline parameter-efficient fine-tuning methods and attention-modulation strategies (such as SoFA, LoRA, and DARA) yield marginal improvements, they fail to fundamentally bridge the inductive correctness gap. In contrast, MMInduction achieves substantial and consistent performance gains across all three base models (Qwen2-VL-7B, Qwen2.5-VL-7B, and Qwen3-VL-8B) under both 1-shot and 3-shot settings. Notably, on complex logical reasoning benchmarks like MMIQ and MME-R, MMInduction not only completely reverses the negative shot effect observed in vanilla models but also establishes new state-of-the-art performance. These comprehensive results compellingly demonstrate that equipping VLMs with an explicit inductive-deductive Chain-of-Thought, supported by balanced visual representations, universally unlocks their multimodal in-context learning potential across tasks of varying cognitive difficulty.

\begin{table}[htbp]
\centering
\caption{Impact of retrieval strategy and multimodal embedding model on downstream ICL performance. Random retrieval consistently degrades accuracy, while Qwen3VL-Emb provides the most stable and effective context matching.}
\label{tab:recall}
\begin{adjustbox}{max width=\textwidth}
\begin{tabular}{lcccccc}
\toprule
\multirow{2}{*}{\diagbox[width=13em, height=2.5\line]{\textbf{Retrieval Setting}}{\textbf{Benchmark}}}
  & \multicolumn{2}{c}{\textbf{VQAv2}}
  & \multicolumn{2}{c}{\textbf{MMIQ}}
  & \multicolumn{2}{c}{\textbf{MDK12}} \\
\cmidrule(lr){2-3} \cmidrule(lr){4-5} \cmidrule(lr){6-7}
  & 1-shot & 3-shot
  & 1-shot & 3-shot
  & 1-shot & 3-shot \\
\midrule
\multicolumn{7}{c}{\textit{\textbf{GPT-5.4}}} \\
\midrule
Zero-shot  & \multicolumn{2}{c}{83.76} & \multicolumn{2}{c}{\underline{37.38}\,/\,\textbf{37.38}} & \multicolumn{2}{c}{57.53} \\
\cdashline{1-7}
Random & 86.89 & 84.89 & 33.59 & 34.88 & 59.01 & 61.67 \\
LlaVE \cite{lan2025llave} & 85.90 & 86.67 & 35.60 & 35.30 & 62.23 & 61.59 \\
QQMM-Emb \cite{xue2025improve} & \textbf{88.09} & \textbf{88.83} & \textbf{38.36} & \underline{36.68} & \underline{62.31} & \textbf{64.00} \\
Qwen3VL-Emb \cite{li2026qwen3} & \underline{87.39} & \underline{88.00} & 37.11 & 36.46 & \textbf{63.17} & \underline{63.21} \\
\midrule
\multicolumn{7}{c}{\textit{\textbf{Claude-4.6-Sonnet}}} \\
\midrule
Zero-shot  & \multicolumn{2}{c}{85.33} & \multicolumn{2}{c}{30.43} & \multicolumn{2}{c}{62.22} \\
\cdashline{1-7}
Random & 85.90 & 86.04 & 30.06 & 32.47 & 61.10 & 63.69 \\
LlaVE \cite{lan2025llave} & 86.13 & 87.56 & 32.76 & 31.69 & 61.88 & 65.08 \\
QQMM-Emb \cite{xue2025improve} & \underline{86.85} & \textbf{90.05} & \textbf{33.18} & \underline{33.03} & \textbf{65.43} & \textbf{66.92} \\
Qwen3VL-Emb \cite{li2026qwen3} & \textbf{87.61} & \underline{89.60} & \underline{32.99} & \textbf{33.29} & \underline{64.64} & \underline{66.44} \\
\midrule
\multicolumn{7}{c}{\textit{\textbf{Qwen3VL-8B-Instruct}}} \\
\midrule
Zero-shot  & \multicolumn{2}{c}{83.62} & \multicolumn{2}{c}{32.30} & \multicolumn{2}{c}{46.59} \\
\cdashline{1-7}
Random & 86.32 & 85.90 & 27.78 & 25.20 & 44.38 & 47.27 \\
LlaVE \cite{lan2025llave} & 86.38 & 88.34 & 27.30 & 27.02 & 46.38 & 48.12 \\
QQMM-Emb \cite{xue2025improve} & \underline{87.43} & 89.30 & \underline{29.65} & \underline{27.53} & \underline{48.87} & \underline{49.40} \\
Qwen3VL-Emb \cite{li2026qwen3} & \textbf{87.75} & \textbf{89.74} & 28.70 & 27.22 & 48.47 & 48.98 \\
Qwen3VL-Emb \& MMInduction & 86.88 & \underline{89.46} & \textbf{35.73} & \textbf{38.46} & \textbf{54.90} & \textbf{55.51} \\
\bottomrule
\end{tabular}
\end{adjustbox}
\end{table}

Table \ref{tab:recall} delineates the impact of various retrieval strategies and embedding models on the downstream performance of multimodal ICL. The empirical results reveal that employing a random retrieval strategy consistently degrades model performance compared to the zero-shot baseline across multiple evaluators (e.g., GPT-5.4, Claude-4.6-Sonnet, and Qwen3VL-8B). This degradation highlights the vulnerability of existing VLMs to noisy or irrelevant contexts and emphasizes the critical necessity of semantic relevance in ICL demonstrations. When evaluating different multimodal embedding models (LlaVE, QQMM-Emb, and Qwen3VL-Emb), we observe that their representational capabilities for image-text pairs are generally comparable, yielding largely similar retrieved cases and downstream accuracies without statistically significant disparities. However, Qwen3VL-Emb demonstrates highly stable and slightly superior heuristic context matching, particularly when paired with the Qwen-VL family as the backbone generator. Consequently, the integration of Qwen3VL-Emb with our MMInduction framework ensures a high-quality, noise-minimized reference context, providing a reliable foundation for the subsequent inductive-deductive reasoning process.

\begin{figure}[htbp]
    \centering
    \begin{subfigure}[b]{0.32\textwidth}
        \centering
        \includegraphics[width=\textwidth]{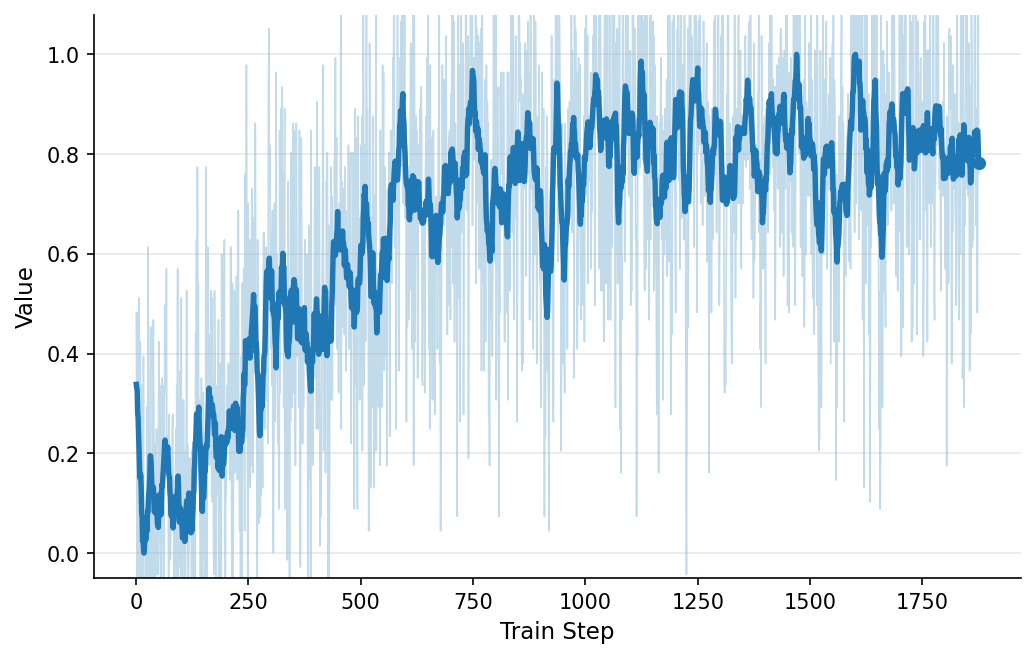}
        \caption{Helpful Reward}
        \label{fig:sub1}
    \end{subfigure}
    \hfill
    \begin{subfigure}[b]{0.32\textwidth}
        \centering
        \includegraphics[width=\textwidth]{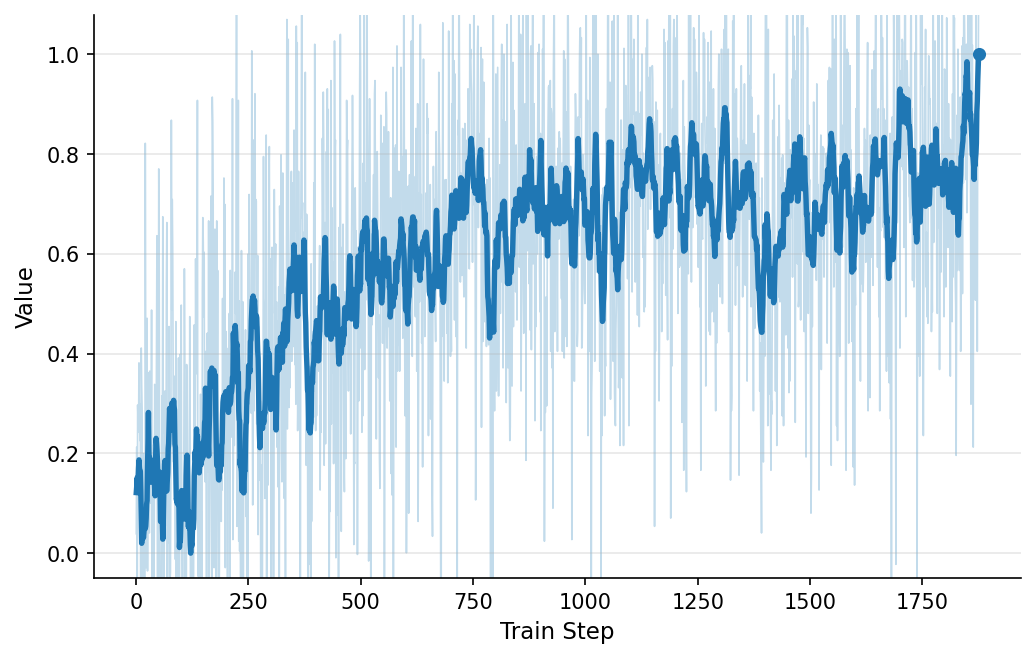}
        \caption{Cite Reward}
        \label{fig:sub2}
    \end{subfigure}
    \hfill
    \begin{subfigure}[b]{0.32\textwidth}
        \centering
        \includegraphics[width=\textwidth]{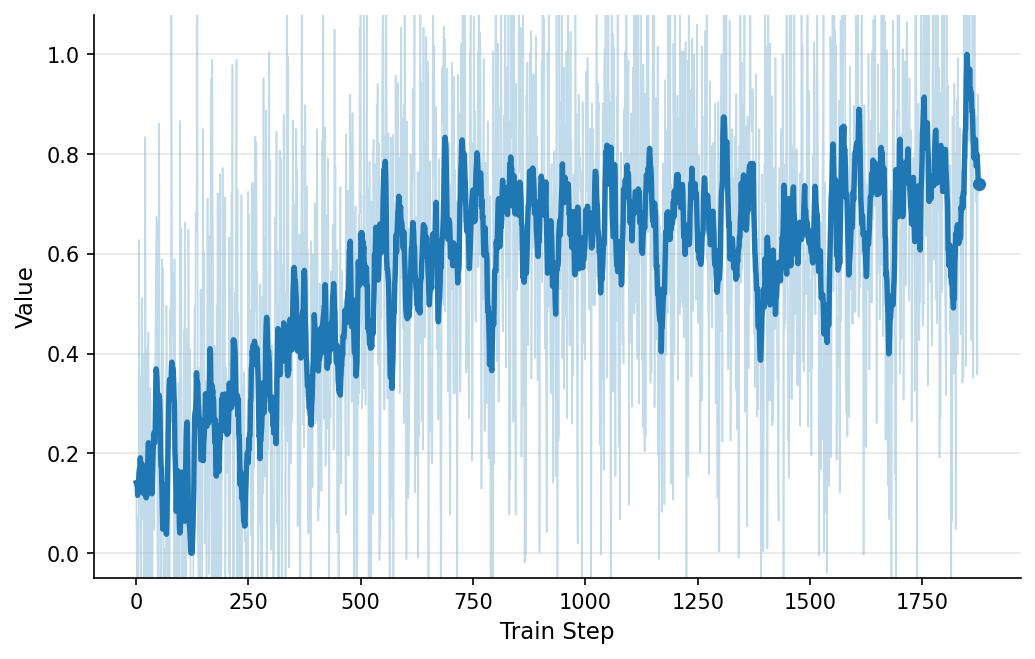}
        \caption{Task Reward}
        \label{fig:sub3}
    \end{subfigure}
    \caption{Training dynamics of the three fine-grained reward components—$R_{\text{task}}$, $R_{\text{helpful}}$, and $R_{\text{cite}}$—during the RLVR phase under the DAPO algorithm.}
    \label{fig:reward}
\end{figure}

Figure \ref{fig:reward} illustrates the evolutionary trajectories of three fine-grained reward components—namely, the task reward ($R_{\text{task}}$), the helpfulness identification reward ($R_{\text{helpful}}$), and the citation compliance reward ($R_{\text{cite}}$)—throughout the reinforcement learning phase. As the training steps progress, all three reward metrics exhibit a robust and monotonic increasing trend, eventually converging toward their respective theoretical upper bounds. Specifically, the rapid convergence of $R_{\text{helpful}}$ and $R_{\text{cite}}$ indicates that the model swiftly acquires the auxiliary capabilities to accurately distinguish noisy distractors from helpful cases and to rigorously cite these cases during its reasoning process. Concurrently, the steady ascent of the primary objective, $R_{\text{task}}$, confirms that the model's final answer accuracy is progressively enhanced without suffering from reward hacking. This synchronous optimization trajectory validates the efficacy of our DAPO-based RLVR pipeline, proving that the carefully designed composite reward function $R = \alpha R_{\text{task}} + \beta R_{\text{helpful}} + \gamma R_{\text{cite}}$ successfully aligns the model's generation policy with the strict requirements of structured inductive-deductive reasoning.

\begin{figure}[htbp]
\centering
\setlength{\tabcolsep}{4pt}
\renewcommand{\arraystretch}{1.1}

\newcommand{\imgcell}[2]{%
    \begin{tabular}{@{}c@{}}
        \includegraphics[width=0.22\textwidth, height=0.16\textwidth]{#1} \\
        \small #2
    \end{tabular}%
}

\begin{tabular}{cccc}
\imgcell{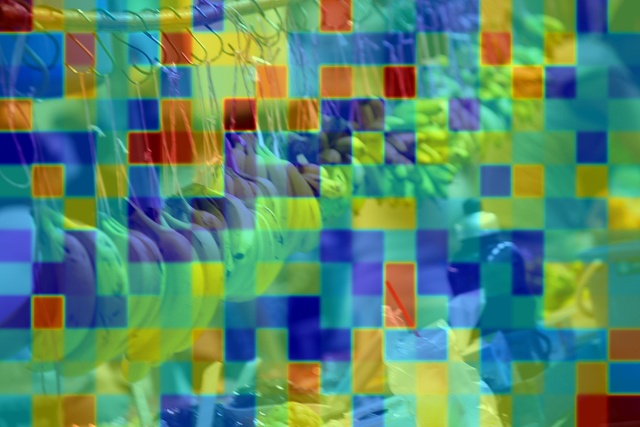}{VQA Case 1} &
\imgcell{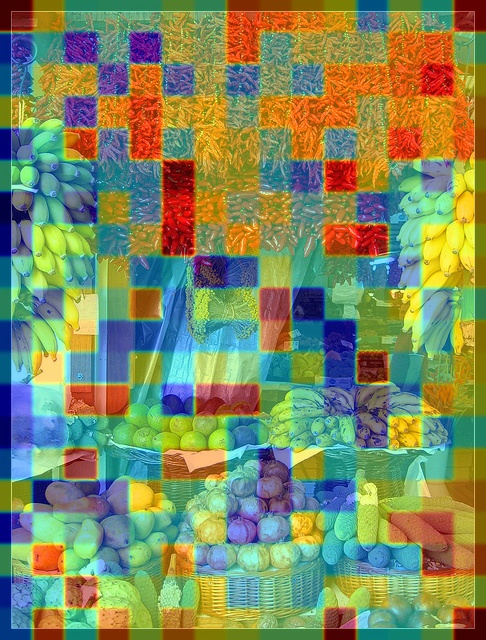}{VQA Case 2} &
\imgcell{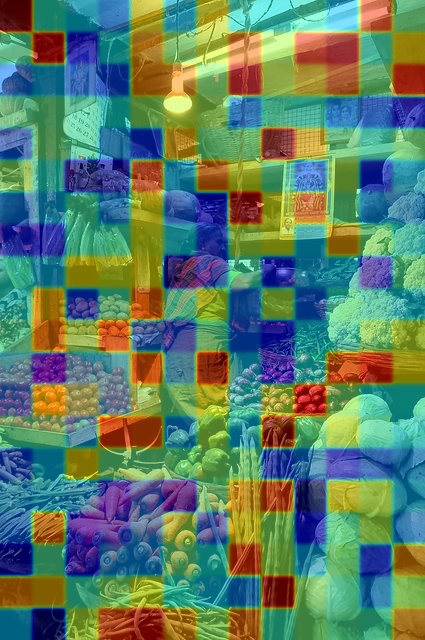}{VQA Case 3} &
\imgcell{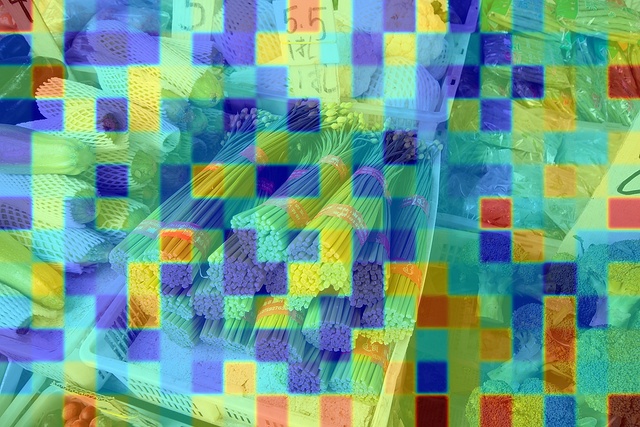}{VQA Query} \\[4pt]
\imgcell{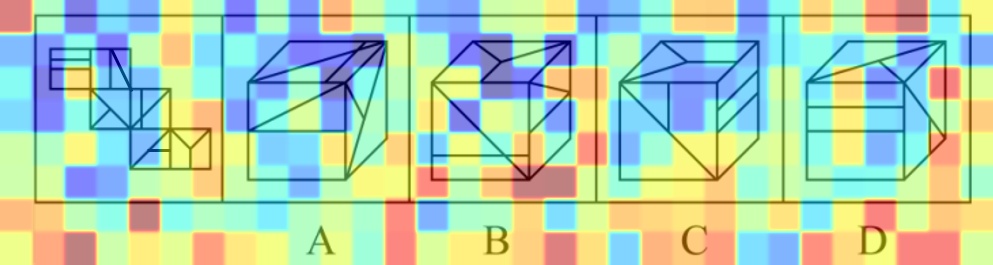}{MMIQ Case 1} &
\imgcell{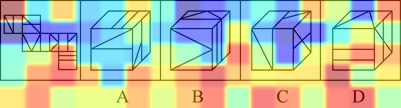}{MMIQ Case 2} &
\imgcell{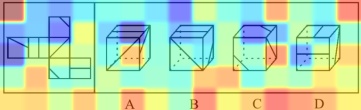}{MMIQ Case 3} &
\imgcell{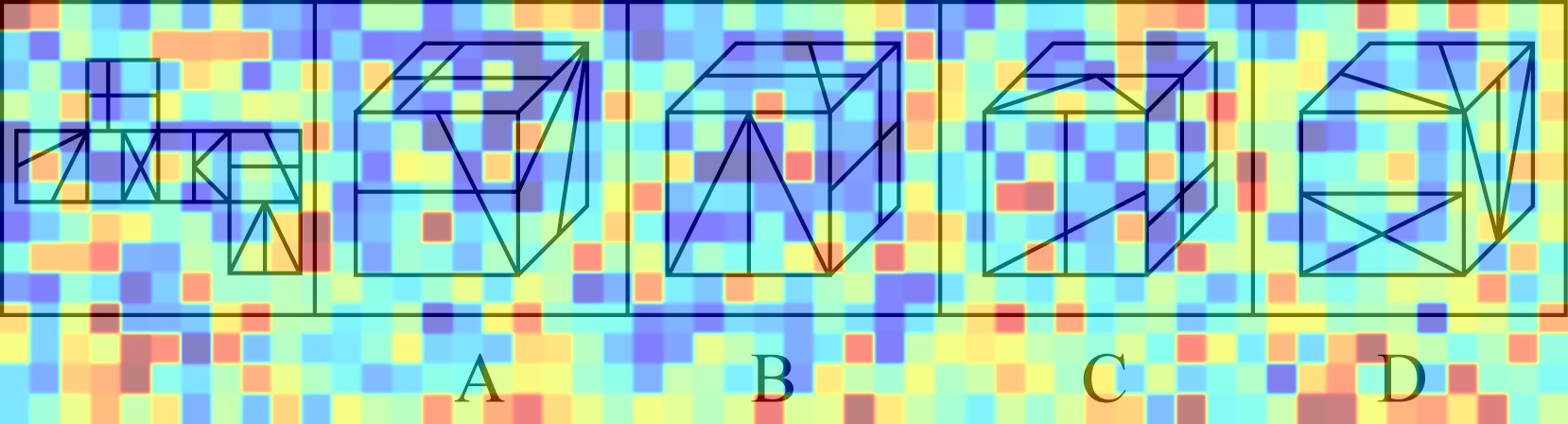}{MMIQ Query} \\[4pt]
\imgcell{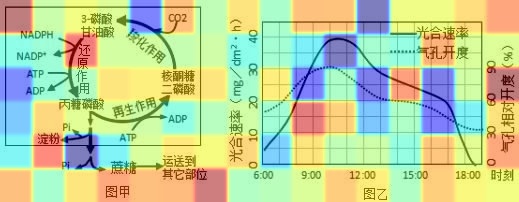}{MDK12 Case 1} &
\imgcell{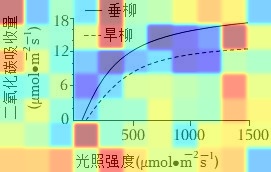}{MDK12 Case 2} &
\imgcell{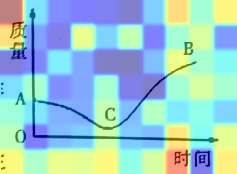}{MDK12 Case 3} &
\imgcell{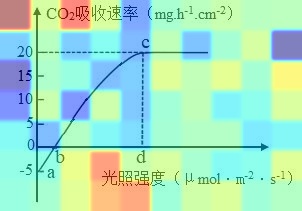}{MDK12 Query} \\[4pt]
\imgcell{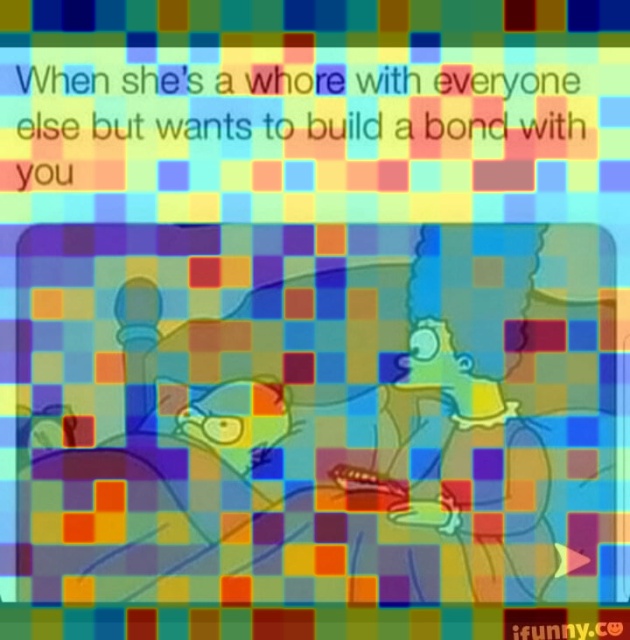}{MAMI Case 1} &
\imgcell{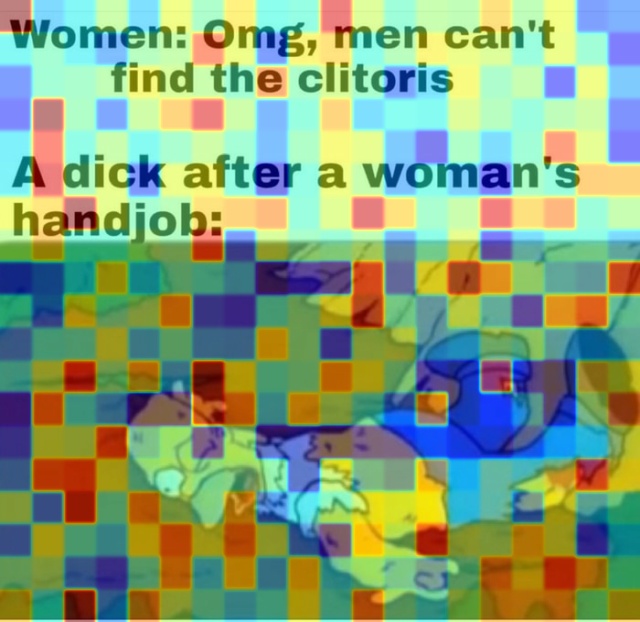}{MAMI Case 2} &
\imgcell{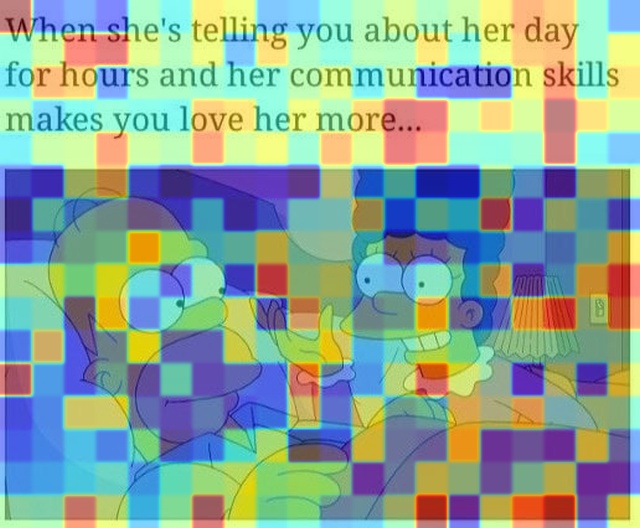}{MAMI Case 3} &
\imgcell{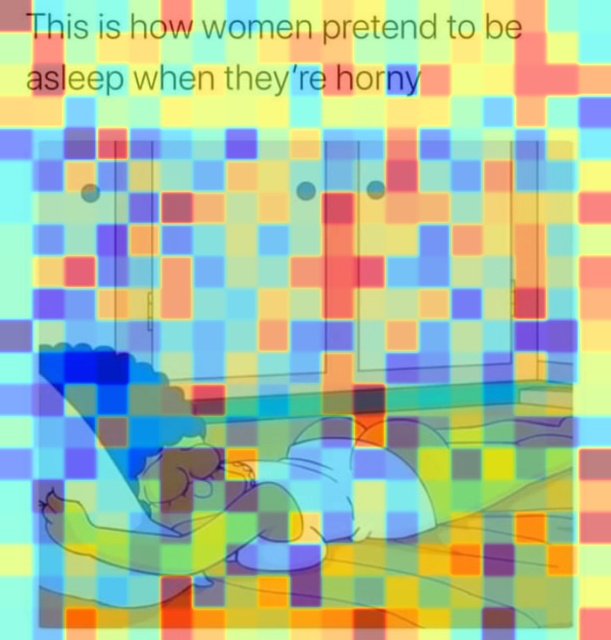}{MAMI Query} \\
\end{tabular}
\caption{Attention heatmaps of VLMs after MMInduction training, demonstrating a balanced attention distribution across all in-context images and the query, in contrast to the first-image bias observed in vanilla models.}
\label{fig:heatmap_new}
\end{figure}

To elucidate the efficacy of the proposed visual attention allocation mechanism, Figure \ref{fig:heatmap_new} provides a comparative visualization of the attention heatmaps generated after MMInduction training. As identified in our motivation study, vanilla VLMs suffer from a severe "attention imbalance," where the model's focus disproportionately collapses onto the first in-context image, leaving subsequent reference cases and the target query severely underutilized. The updated heatmaps in Figure 6 demonstrate that our dynamic visual attention module successfully mitigates this first-image bias. Following the adaptive reweighting of the attention score matrix, the model exhibits a highly equitable and balanced attention distribution across the entire multimodal context sequence. Sufficient computational focus is now allocated to intermediate cases and the final query image, ensuring that key visual semantics from all demonstrations are actively processed. This balanced sensory foundation proves that MMInduction effectively dismantles representation-level barriers, enabling the model to perform holistic cross-case pattern discovery which is an absolute prerequisite for genuine inductive reasoning.

\section{Detailed Experimental Settings}
\label{app:training}

All experiments were conducted on a workstation equipped with 8×NVIDIA A100 GPUs running Ubuntu 24.04.2 LTS, using PyTorch 2.6.0 with CUDA 12.9.  In the experiments, we conducted inference and training of open-source models based on the MS-Swift \cite{zhao2025swift} framework. For inference, the default parameters of the framework were adopted. The parameters used during SFT and RLVR are shown in table \ref{tab:sft-rl-params}. Code, data and partial results of the experiment are all provided in the supplementary materials.

\begin{table}[h]
\centering
\caption{Training Parameters for SFT and RL}
\begin{minipage}[t]{0.48\columnwidth}
\centering
\caption*{SFT}
\begin{adjustbox}{max width=\columnwidth}
\begin{tabular}{c|c}
\toprule
\textbf{Parameter} & \textbf{Value} \\
\midrule
num\_train\_epochs & 3 \\
train\_type & lora \\
lora\_rank & 8 \\
freeze\_vit & True \\
freeze\_adapter & True \\
attn\_impl & sdpa \\
max\_pixels 1505280 \\
torch\_dtype & bfloat16 \\
learning\_rate & $1 \times 10^{-5}$ \\
warmup\_ratio & 0.05 \\
target\_modules & all-linear \\
per\_device\_train\_batch\_size & 4 \\
gradient\_accumulation\_steps & 4 \\
use\_liger\_kernel & True \\
\bottomrule
\end{tabular}
\end{adjustbox}
\end{minipage}
\hfill
\begin{minipage}[t]{0.48\columnwidth}
\centering
\caption*{RL}
\begin{adjustbox}{max width=\columnwidth}
\begin{tabular}{c|c}
\toprule
\textbf{Parameter} & \textbf{Value} \\
\midrule
num\_train\_epochs & 1 \\
train\_type & lora \\
torch\_dtype & bfloat16 \\
max\_length & 10240 \\
max\_completion\_length & 2048 \\
num\_generations & 8 \\
learning\_rate & $1 \times 10^{-5}$ \\
warmup\_ratio & 0.05 \\
target\_modules & all-linear \\
per\_device\_train\_batch\_size & 1 \\
gradient\_accumulation\_steps & 8 \\
temperature & 1.0 \\
top\_p & 0.85 \\
dynamic\_sample & True \\
overlong\_filter & True \\
\bottomrule
\end{tabular}
\end{adjustbox}
\end{minipage}
\label{tab:sft-rl-params}
\end{table}

The construction of the SFT dataset involved a rigorous, multi-step distillation and verification pipeline. Rather than generating the reasoning chain end-to-end, we prompted Gemini-3.1-pro \cite{pichai2025gemini3} to sequentially generate the case analysis, rule induction, and final deduction steps. To guarantee the logical soundness of the synthesized data, the generated reasoning chains were subjected to a voting mechanism by a panel of three expert models: GPT-5.4 \cite{singh2025openai}, Claude-4.6-Sonnet \cite{anthropic2025claude}, and Seed-1.8-pro \cite{seed2026seed1}. A synthesized sample was retained for training only if it secured at least a two-thirds majority agreement regarding its logical validity. Furthermore, to enhance the model's robustness against irrelevant context, hard negative cases were introduced into the prompts. These noise cases were constructed by computing the similarity scores across the retrieval database, isolating the bottom-$k$ samples with the lowest similarity to the query, and subsequently performing random sampling from this subset.

For the evaluation phase, we employed distinct LLM-as-a-Judge protocols tailored to different metrics, with the API temperature strictly set to $0$ to ensure deterministic and stable outputs. The assessment of Answer Accuracy was conducted solely by GPT-5.4-mini \cite{singh2025openai}. Given that the ground truth labels for our datasets are predominantly standard formats (e.g., true/false, multiple-choice, and fill-in-the-blank), this evaluation requires minimal complex reasoning, making a single high-performance LLM highly reliable. Conversely, evaluating Induction Accuracy necessitates a deeper understanding of reasoning patterns. For this metric, the ground truth rules were manually annotated by human experts. The evaluation was then performed via a majority vote among three models: GPT-5.4-mini \cite{singh2025openai}, Gemini-3.0-flash \cite{pichai2025gemini3}, and Claude-4.5-haiku \cite{anthropic2025claude}. To validate the reliability of this LLM judge ensemble, two human experts randomly sampled and cross-examined 150 evaluated instances each. The inter-rater agreement between the LLM ensemble and human experts achieved a Cohen's Kappa \cite{sun2011meta} coefficient of $0.78$, demonstrating a high degree of alignment and confirming the validity of our automated evaluation pipeline.

\section{Dataset Details}
\label{app:datasets}

\paragraph{\textbf{VQAv2} \cite{antol2015vqa}}  is a large-scale visual question answering benchmark. The dataset is built upon images from the Microsoft COCO dataset and contains open-ended questions that require joint understanding of visual content and natural language. The original dataset comprises 443,757 training samples and 214,354 validation samples, with each question annotated with 10 ground truth answers to account for answer variability. In our experiments, we sampled 2,000 instances from the original validation set as our training data and 234 instances as our test set. During the retrieval phase, we utilized the complete validation set containing 214,354 samples as the knowledge base to ensure comprehensive coverage for similarity-based retrieval operations.

\paragraph{\textbf{TextVQA} \cite{singh2019towards}} is a reading-comprehension-oriented visual question answering dataset. Unlike conventional VQA datasets that primarily focus on visual object recognition and spatial reasoning, TextVQA specifically requires models to read and reason about text present within images to answer questions correctly. The dataset contains images with rich textual content such as signs, labels, and documents, presenting unique challenges that combine optical character recognition with visual reasoning. The original dataset consists of 34,602 training samples and 5,000 validation samples. For our experiments, we sampled 2,000 instances from the original training set as our training data and 164 instances from the original validation set as our test set.

\paragraph{\textbf{MMIQ} \cite{cai2025mm}} is a comprehensive evaluation framework designed to benchmark human-like abstraction and reasoning in multimodal models. The benchmark functions as a visual IQ test and comprises 2,710 meticulously curated test items spanning 8 distinct reasoning paradigms, including Logical Operation, 2D-Geometry, 3D-Geometry, Visual Instruction, Temporal Movement, Spatial Relationship, Concrete Object, and Mathematical reasoning. The dataset aims to isolate core cognitive competencies by decoupling assessment from linguistic background or domain-specific knowledge, utilizing visual puzzles with simple 2D and 3D shapes to evaluate pattern recognition, analogical transfer, discrimination, and generalization abilities. From the total of 2,710 samples, we sampled 387 instances as our test set and used the remaining 2,323 instances as our training data.

\paragraph{\textbf{MME-Reasoning} \cite{yuan2025mme}} is a comprehensive benchmark specifically designed to evaluate the logical reasoning capability of Multimodal Large Language Models. The benchmark consists of 1,188 curated questions covering three fundamental types of logical reasoning: inductive, deductive, and abductive reasoning, spanning various difficulty levels. The dataset is carefully constructed to ensure that questions evaluate reasoning ability specifically rather than perceptual skills or general knowledge breadth, providing a focused assessment of logical inference capabilities in multimodal contexts. From the total of 1,188 samples, we sampled 200 instances as our test set and used the remaining 988 instances as our training data.

\paragraph{\textbf{MDK12} \cite{zhou2026mdk12}} is a multi-discipline benchmark for evaluating reasoning capabilities in multimodal large language models. The benchmark is constructed from real-world K-12 educational examination questions spanning multiple academic disciplines, providing a comprehensive assessment of models' abilities to understand and reason about educational content across different subject areas and grade levels. The dataset covers diverse topics including mathematics, science, and language arts, requiring both visual comprehension and domain-specific reasoning. After data cleaning and quality filtering, we obtained 1,811 usable samples. From these, we sampled 330 instances as our test set and used the remaining 1,481 instances as our training data.

\paragraph{\textbf{CSVQA} \cite{jian2025csvqa}} is a Chinese multimodal benchmark specifically designed for evaluating STEM reasoning capabilities of vision-language models. The benchmark focuses on scientific reasoning through domain-grounded visual question answering, featuring questions that require understanding of scientific diagrams, mathematical figures, and technical illustrations commonly encountered in Chinese STEM education contexts. The dataset presents unique challenges by combining Chinese language understanding with visual scientific reasoning across physics, chemistry, biology, and mathematics domains. The original dataset contains 1,378 samples, from which we sampled 195 instances as our test set and used the remaining 1,183 instances as our training data.

 \paragraph{\textbf{HarMeme} \cite{pramanick2021detecting}} is a benchmark dataset for detecting harmful memes and identifying their targets. The dataset contains internet memes related to COVID-19 that have been annotated through a rigorous two-stage process, where memes are first labeled as very harmful, partially harmful, or harmless, and then further annotated with target categories including individual, organization, community, or society and general public. The benchmark addresses the challenge of understanding satirical and contextually dependent multimodal content that requires joint reasoning over visual and textual elements to detect potential harm. The original dataset comprises 4,793 samples, from which we sampled 2,000 instances as our training data and 200 instances as our test set.

 \paragraph{\textbf{MAMI} \cite{fersini2022semeval}} is a multimedia automatic misogyny identification benchmark. The dataset explores the detection of misogynous content in internet memes by leveraging both textual and visual modalities. The benchmark was organized into two related sub-tasks: recognizing whether a meme is misogynous and identifying specific types of misogyny present in harmful memes. The dataset attracted significant attention from the research community with over 400 participants during the shared task, highlighting the importance and challenge of this multimodal content moderation problem. From the available samples, we sampled 2,000 instances as our training data and 200 instances as our test set.

In this work, to ensure balanced distribution across different data categories, we maintained the training data size for each dataset at approximately 2,000 samples, although some datasets contain fewer than 2,000 total instances and thus contributed their full available training portions. To control evaluation time and computational costs, we sampled approximately 200 test instances per dataset. For retrieval operations, we utilized the original complete datasets as knowledge bases whenever possible to maximize the coverage and diversity of retrievable examples. During the sampling process, we enforced a constraint that each selected sample must be able to match at least 5 distinct cases with similarity scores above 80 percent and at least 5 cases with similarity scores below 20 percent from the knowledge base, ensuring sufficient positive and negative examples for similarity-based retrieval and random noise injection during both training and evaluation phases.

\section{Prompt Template}
\label{app:prompt}

\begin{table}[htbp]
\caption{The prompt template for motivation study.}
\resizebox{\linewidth}{!}{
\begin{tcolorbox}
\small
You are an expert in visual tasks. The user will provide you with a question and its reference cases. Please answer the user's question by referring to the cases provided.
Output the thinking process in <think> </think> and  final answer in <answer> </answer> tags.
        
\end{tcolorbox}}
\label{tab:prompt_motivation}
\end{table}

\begin{table}[htbp]
\caption{The prompt template for Inductive CoT.}
\resizebox{\linewidth}{!}{
\begin{tcolorbox}
\small
You are an expert in visual reasoning and problem-solving. The user will provide you with a Target Question (with image) and several Reference Cases. Your goal is to filter out irrelevant noise, extract problem-solving rules from helpful cases, and solve the Target Question.

Please follow these strict steps:
\begin{enumerate}
    \item Target Problem Analysis: First, carefully observe the Target Question and its image. Identify the core task, key visual elements, and the implicit logic required to solve it.
    \item Reference Case Evaluation: Analyze each reference case one by one. For each case, you must explicitly evaluate:
    \begin{itemize}
        \item Visual \& Textual Elements: What is shown in the image and asked in the text?
        \item Solution Logic: What is the underlying rule, formula, or method used to solve this case?
        \item Relevance Comparison: Compare its task type and solution logic with the Target Problem. (Note: Superficial visual similarity does not mean it's helpful; focus on logical and task relevance).
    \end{itemize}
    \item Rule Induction: Synthesize a general problem-solving rule ONLY from the cases judged as "helpful". Ignore "unhelpful" cases completely. Use tags like "\texttt{<|case 1|>}" to cite them.
    \item Final Answer: Apply the induced rule to solve the Target Question step-by-step.
\end{enumerate}

Use the following strict template for your output:

\texttt{<|begin\_of\_target\_analysis|>}\\
\texttt{[Core Task]: \{What is the user asking?\}}\\
\texttt{[Key Visuals]: \{What are the crucial elements in the target image?\}}\\
\texttt{[Required Logic]: \{What kind of reasoning or method is likely needed?\}}\\
\texttt{<|end\_of\_target\_analysis|>}\\

\texttt{<|begin\_of\_reference\_analysis|>}\\
\texttt{<|begin\_of\_case\_1\_analysis|>}\\
\texttt{[Elements]: \{Describe image and text\}}\\
\texttt{[Solution Logic]: \{Explain how the case was solved\}}\\
\texttt{[Comparison]: \{Compare with Target Problem's Core Task and Required Logic\}}\\
\texttt{<|begin\_of\_helpful\_judgment|>helpful (or unhelpful)<|end\_of\_helpful\_judgment|>}\\
\texttt{<|end\_of\_case\_1\_analysis|>}\\

\texttt{<|begin\_of\_case\_2\_analysis|>}\\
\texttt{... (Repeat for all cases)}\\
\texttt{<|end\_of\_case\_2\_analysis|>}\\
\texttt{<|end\_of\_reference\_analysis|>}\\

\texttt{<|begin\_of\_induction|>}\\
\texttt{[Helpful Cases Cited]: \{e.g., <|case\_1|>, <|case\_3|>\}}\\
\texttt{[Extracted Rule]: \{Clearly state the general formula, pattern, or logical rule derived from the helpful cases. This should be a reusable principle that can be directly applied to the Target Question.\}}\\
\texttt{<|end\_of\_induction|>}\\

\texttt{<|begin\_of\_answer|>}\\
\texttt{[Step-by-step Application]: \{Apply the [Extracted Rule] to the Target Problem\}}\\
\texttt{[Final Result]: \{The final answer\}}\\
\texttt{<|end\_of\_answer|>}
        
\end{tcolorbox}}
\label{tab:prompt_induction}
\end{table}

\begin{table}[htbp]
\caption{The prompt template for LLM Accuracy Judge.}
\resizebox{\linewidth}{!}{
\begin{tcolorbox}
\small
\section*{System Role}
You are a rigorous grader.\\
Given (1) the problem, (2) the standard answer, and (3) the student's response, evaluate whether the student's response is correct.

\section*{General Principles}
\begin{itemize}
    \item Treat the standard answer as the ground truth for grading.
    \item Default to strict, unambiguous correctness. If the problem allows multiple correct answers or methods, accept any that are mathematically or logically equivalent to the standard answer and satisfy all stated constraints.
    \item When the problem requires a single final answer, the student must provide exactly one; multiple conflicting answers count as Wrong unless the student clearly indicates a final revised answer.
    \item If the student leaves the answer blank, says "I don't know," or gives only restatement of the problem, all sub-answers are Wrong.
\end{itemize}

\section*{Scoring Method}
\begin{enumerate}
    \item Decompose the standard answer into individual sub-answers (e.g., for a single answer: one sub-answer; for a list/sequence: each element is a sub-answer).
    \item For each sub-answer, determine independently whether the student's response contains a correct match (Correct or Wrong).
    \item The final score = number of correct sub-answers / total number of sub-answers.
\end{enumerate}

\section*{Output Format}
Use the following XML tags:\\

\texttt{<sub\_answers>}\\
\texttt{List each sub-answer on a separate line in the format: [sub-answer content]: Correct or Wrong}\\
\texttt{</sub\_answers>}\\
\texttt{<reason>1--3 concise sentences summarizing the grading result.</reason>}
        
\end{tcolorbox}}
\label{tab:prompt_acc_judge}
\end{table}

\begin{table}[htbp]
\caption{The prompt template for LLM Induction Judge.}
\resizebox{\linewidth}{!}{
\begin{tcolorbox}
\small
\section*{System Role}
You are an expert assessor in reasoning patterns.\\
Given (1) the problem (a pattern recognition task), (2) the expert-derived reasoning pattern, and (3) the student's proposed reasoning pattern, your task is to determine whether the student's reasoning pattern is \textbf{basically consistent} with the expert's pattern.

\section*{Assessment Principles}
\begin{itemize}
    \item \textbf{Focus on conceptual alignment, not wording.} The student's reasoning need not match the expert's verbatim, but must capture the same underlying logical or mathematical relationship.
    \item \textbf{Allow equivalent formulations.} If the student describes the same pattern using different terms, examples, or representations, accept it as consistent.
    \item \textbf{Reject fundamentally different or incomplete reasoning.} If the student's pattern misses a key step, misidentifies the core relationship, or introduces unrelated logic, mark as inconsistent.
    \item \textbf{Be conservative with partial matches.} If the student's reasoning includes only part of the expert pattern or conflates multiple patterns, mark as inconsistent unless the core mechanism is fully and correctly captured.
    \item \textbf{Ignore calculation errors or final-answer mistakes.} Your judgment is based \textbf{only} on the reasoning pattern described, not on numerical accuracy or the final answer given.
\end{itemize}

\section*{Output Format}
Use the following XML tags in your response:\\

\texttt{<reason>1--4 concise sentences explaining why the student's reasoning does or does not align with the expert's</reason>}\\
\texttt{<pattern\_match>Consistent or Inconsistent</pattern\_match>}
        
\end{tcolorbox}}
\label{tab:prompt_induction_judge}
\end{table}

Table \ref{tab:prompt_motivation} illustrates the prompt designed to evaluate the baseline inductive capabilities of existing VLMs during our preliminary motivation study. By explicitly instructing the model to output its thinking process before generating the final answer, this prompt allows us to dissect whether the model genuinely extracts underlying rules from the provided cases or merely relies on shallow pattern matching. This setup is crucial for exposing the "inductive correctness gap" that fundamentally limits current multimodal in-context learning.

Table \ref{tab:prompt_induction} presents the comprehensive prompt template for our proposed Inductive CoT. This prompt structurally enforces a rigorous reasoning pipeline, guiding the model to first analyze the target problem, systematically evaluate each reference case to filter out unhelpful noise, explicitly synthesize a generalizable rule, and finally deduce the answer step-by-step. This structured guidance is the cornerstone of MMInduction, transforming multimodal ICL from naive jump-to-answer behaviors into genuine logical reasoning.

Table \ref{tab:prompt_acc_judge} details the prompt utilized to instruct the LLM judge in evaluating the final answer correctness. To ensure rigorous and objective grading, the prompt establishes strict general principles, such as treating the standard answer as the absolute ground truth and requiring exact matches for single-answer questions. By decomposing complex answers into sub-answers and scoring them independently, this prompt guarantees a highly reliable and reproducible automated evaluation metric across diverse datasets.

Table \ref{tab:prompt_induction_judge} shows the prompt designed for assessing the induction accuracy of the models. Unlike standard accuracy evaluation, this prompt specifically directs the judge to focus on conceptual alignment and the underlying logical or mathematical relationships, explicitly ignoring superficial wording differences or final calculation errors. This nuanced evaluation framework is essential for accurately measuring whether a model has successfully acquired the intrinsic problem-solving rule from the context.

\section{Case Study}
\label{app:case_study}

\newtcolorbox{headerbox}{
  enhanced, colback=gray!15, colframe=gray!60,
  boxrule=0.4pt, arc=2pt,
  left=6pt, right=6pt, top=3pt, bottom=3pt, boxsep=2pt
}
\newtcolorbox{querybox}{
  enhanced, colback=blue!4, colframe=blue!55!black,
  title=\textbf{Query}, fonttitle=\small,
  boxrule=0.5pt, arc=2pt,
  left=5pt, right=5pt, top=3pt, bottom=3pt
}
\newtcolorbox{demobox}{
  enhanced, colback=orange!5, colframe=orange!70!black,
  title=\textbf{Retrieved Cases}, fonttitle=\small,
  boxrule=0.5pt, arc=2pt,
  left=4pt, right=4pt, top=3pt, bottom=3pt
}
\newtcolorbox{wrongbox}[1]{
  enhanced, colback=red!4, colframe=red!60!black,
  title=\textbf{#1}, fonttitle=\small,
  boxrule=0.4pt, arc=2pt,
  left=5pt, right=5pt, top=2pt, bottom=2pt
}
\newtcolorbox{rightbox}[1]{
  enhanced, colback=green!5, colframe=green!45!black,
  title=\textbf{#1}, fonttitle=\small,
  boxrule=0.5pt, arc=2pt,
  left=5pt, right=5pt, top=3pt, bottom=3pt
}

\newcommand{\cmark}{\textcolor{green!55!black}{\ding{51}}}
\newcommand{\xmark}{\textcolor{red!70!black}{\ding{55}}}

\FloatBarrier
\par\addvspace{4pt}

\noindent
\begin{headerbox}
\centering
\textbf{Dataset:} VQA \quad$|$\quad
\textbf{Task:} Multimodal Knowledge Reasoning \quad$|$\quad
\textbf{Case ID:} \#199720001
\end{headerbox}

\vspace{5pt}

\noindent
\begin{minipage}[t]{0.48\linewidth}
\vspace{0pt}
\tcbset{equal height group=casestudy-VQA-199720001}
  \begin{querybox}
    \centering
    \includegraphics[width=0.72\linewidth]{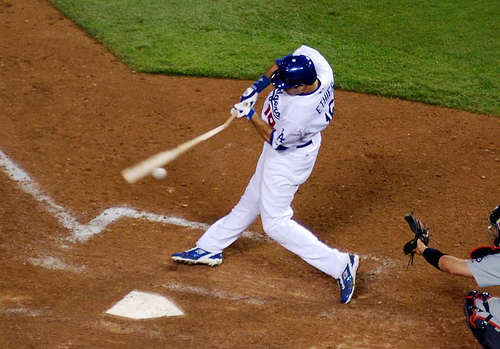}\\[4pt]
    \raggedright\small
    \textbf{Q:} What color is the batter's helmet?\par\vspace{2pt}
    \textbf{GT Answer:} blue
  \end{querybox}
\end{minipage}\hfill
\begin{minipage}[t]{0.50\linewidth}
\vspace{0pt}
\tcbset{equal height group=casestudy-VQA-199720001}
  \begin{demobox}
    \small
    \begin{minipage}[c]{0.26\linewidth}\centering
      \includegraphics[width=\linewidth]{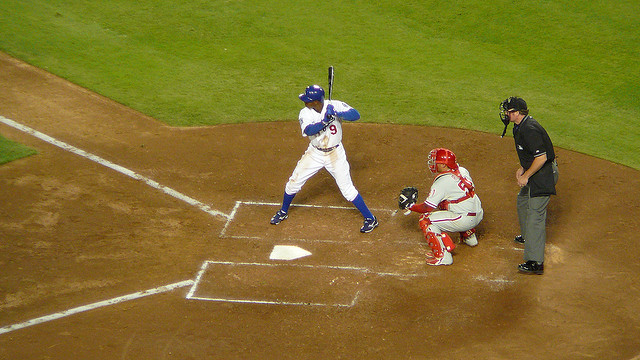}
    \end{minipage}\hfill
    \begin{minipage}[c]{0.70\linewidth}\raggedright
      \textbf{Q1:} What color is the batter's helmet?\par
      \textbf{A1:} blue
    \end{minipage}
    \par\vspace{3pt}\hrule\vspace{3pt}
    \begin{minipage}[c]{0.26\linewidth}\centering
      \includegraphics[width=\linewidth]{case/AccCase/VQA/VQA_Case2.jpg}
    \end{minipage}\hfill
    \begin{minipage}[c]{0.70\linewidth}\raggedright
      \textbf{Q2:} What color is the batter's helmet?\par
      \textbf{A2:} blue
    \end{minipage}
    \par\vspace{3pt}\hrule\vspace{3pt}
    \begin{minipage}[c]{0.26\linewidth}\centering
      \includegraphics[width=\linewidth]{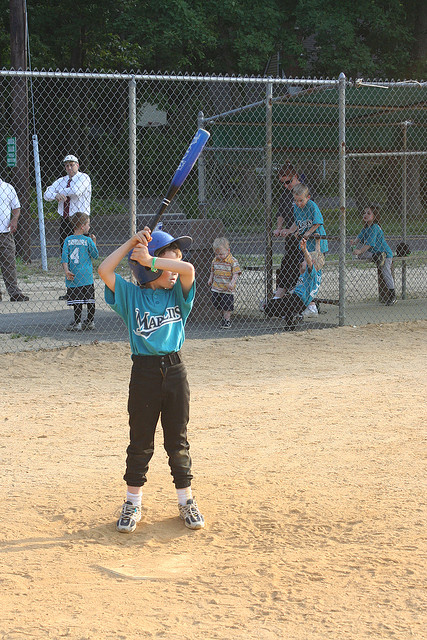}
    \end{minipage}\hfill
    \begin{minipage}[c]{0.70\linewidth}\raggedright
      \textbf{Q3:} What color is the little boy's helmet?\par
      \textbf{A3:} blue
    \end{minipage}
  \end{demobox}
\end{minipage}

\vspace{8pt}

\tcbset{breakable, enforce breakable}

\begin{rightbox}{Vanilla Gemini-3.1-pro \cmark}
  \small
  Based on the image, the color of the batter's helmet is blue.
\end{rightbox}

\vspace{5pt}

\begin{rightbox}{Vanilla Qwen3-VL-8B \cmark}
  \small
  blue
\end{rightbox}

\vspace{5pt}

\begin{rightbox}{Qwen3-VL-8B+MMInduction (Ours) \cmark}
  \scriptsize
  \setlength{\columnsep}{12pt}
  \setlength{\parskip}{2pt}
  \linespread{0.95}\selectfont
  <|begin\_of\_target\_analysis|>\\
{[Core Task]}: Determine the color of the batter's helmet.\\
{[Key Visuals]}: A baseball batter in white uniform swinging at a pitch; the batter is wearing a helmet.\\
{[Required Logic]}: Identify the batter and visually inspect the helmet's color.\\
<|end\_of\_target\_analysis|>

<|begin\_of\_reference\_analysis|>\\
<|begin\_of\_case\_1\_analysis|>\\
{[Elements]}: Baseball batter at home plate; question asks the color of the batter's helmet.\\
{[Solution Logic]}: Locate batter, look at helmet, identify its color as blue.\\
{[Comparison]}: Same sport context (baseball), same object (batter's helmet), same task (determine color).\\
<|begin\_of\_helpful\_judgment|>helpful<|end\_of\_helpful\_judgment|>\\
<|end\_of\_case\_1\_analysis|>

<|begin\_of\_case\_2\_analysis|>\\
{[Elements]}: Batter swinging during a game; question asks helmet color.\\
{[Solution Logic]}: Visual inspection of batter's helmet color.\\
{[Comparison]}: Identical task and logic to target problem.\\
<|begin\_of\_helpful\_judgment|>helpful<|end\_of\_helpful\_judgment|>\\
<|end\_of\_case\_2\_analysis|>

<|begin\_of\_case\_3\_analysis|>\\
{[Elements]}: Young boy holding a bat; question asks color of his helmet.\\
{[Solution Logic]}: Identify helmet and determine its color.\\
{[Comparison]}: Same type of reasoning (color of helmet) though different subject.\\
<|begin\_of\_helpful\_judgment|>helpful<|end\_of\_helpful\_judgment|>\\
<|end\_of\_case\_3\_analysis|>\\
<|end\_of\_reference\_analysis|>

<|begin\_of\_induction|>\\
{[Helpful Cases Cited]}: <|case\_1|>, <|case\_2|>, <|case\_3|>\\
{[Extracted Rule]}: To answer questions about the color of a helmet, first locate the relevant person (batter or child), then visually inspect the helmet on their head and report its dominant color.\\
<|end\_of\_induction|>

<|begin\_of\_answer|>\\
{[Step-by-step Application]}:  \\
- Locate the batter in the target image: he is swinging at the ball near home plate.  \\
- Look at the helmet on his head.  \\
- The helmet appears to be blue.

{[Final Result]}: blue\\
<|end\_of\_answer|>
\end{rightbox}

\vspace{4pt}
{\setlength{\abovecaptionskip}{6pt}%
\captionof{figure}{Case study on the VQA dataset.}%
\label{fig:case_study_vqa}}

\FloatBarrier

\FloatBarrier
\par\addvspace{4pt}

\noindent
\begin{headerbox}
\centering
\textbf{Dataset:} MMIQ \quad$|$\quad
\textbf{Task:} Multimodal Knowledge Reasoning \quad$|$\quad
\textbf{Case ID:} \#364
\end{headerbox}

\vspace{5pt}

\noindent
\begin{minipage}[t]{0.48\linewidth}
\vspace{0pt}
\tcbset{equal height group=casestudy-a_only-MMIQ-364}
  \begin{querybox}
    \centering
    \includegraphics[width=0.72\linewidth]{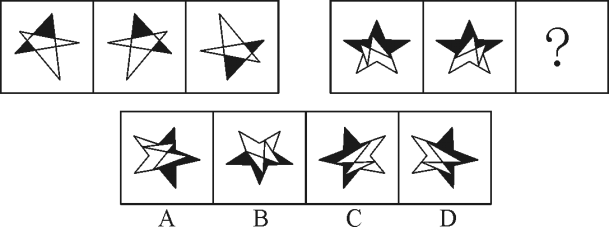}\\[4pt]
    \raggedright\small
    \textbf{Q:} Choose the most appropriate option from the given four choices to fill in the question mark, so that it presents a certain regularity:\par\vspace{2pt}
    \textbf{GT Answer:} B
  \end{querybox}
\end{minipage}\hfill
\begin{minipage}[t]{0.50\linewidth}
\vspace{0pt}
\tcbset{equal height group=casestudy-a_only-MMIQ-364}
  \begin{demobox}
    \small
    \begin{minipage}[c]{0.26\linewidth}\centering
      \includegraphics[width=\linewidth]{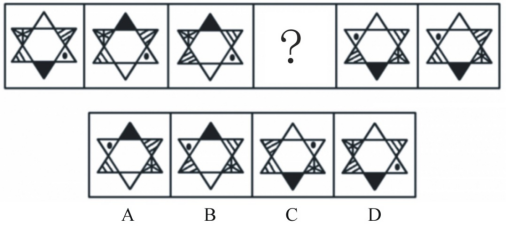}
    \end{minipage}\hfill
    \begin{minipage}[c]{0.70\linewidth}\raggedright
      \textbf{Q1:} Choose the most appropriate option from the given four choices to fill in the question mark, so that it presents a certain regularity:\par
      \textbf{A1:} B
    \end{minipage}
    \par\vspace{3pt}\hrule\vspace{3pt}
    \begin{minipage}[c]{0.26\linewidth}\centering
      \includegraphics[width=\linewidth]{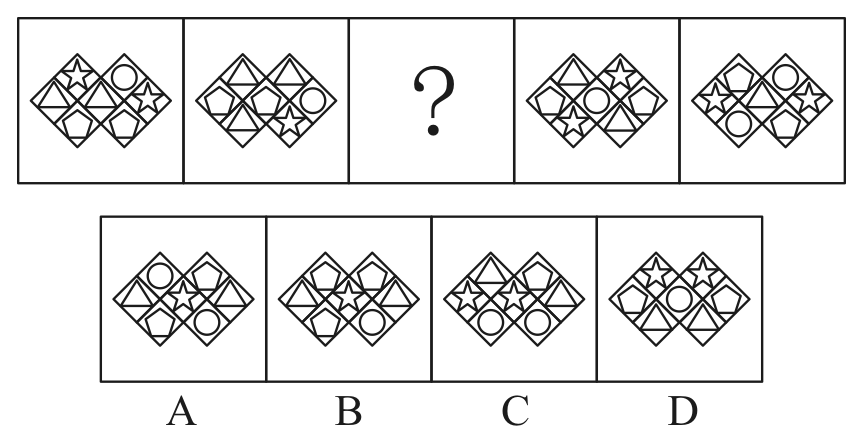}
    \end{minipage}\hfill
    \begin{minipage}[c]{0.70\linewidth}\raggedright
      \textbf{Q2:} Choose the most appropriate option from the given four choices to fill in the question mark, so that it presents a certain regularity:\par
      \textbf{A2:} B
    \end{minipage}
    \par\vspace{3pt}\hrule\vspace{3pt}
    \begin{minipage}[c]{0.26\linewidth}\centering
      \includegraphics[width=\linewidth]{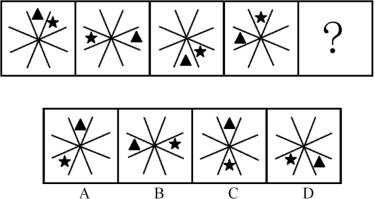}
    \end{minipage}\hfill
    \begin{minipage}[c]{0.70\linewidth}\raggedright
      \textbf{Q3:} Choose the most appropriate option from the given four choices to fill in the question mark, so that it presents a certain regularity:\par
      \textbf{A3:} A
    \end{minipage}
  \end{demobox}
\end{minipage}

\vspace{8pt}

\tcbset{breakable, enforce breakable}

\begin{wrongbox}{Vanilla Gemini-3.1-pro \xmark}
  \small
  Based on an analysis of the visual patterns in the given image:

1.  \textbf{Analyze the First Group of Figures:} \\
    *   The first set of three panels shows a complex star-like shape with certain regions shaded black. \\
    *   If we observe the general position of the "main" shaded mass: \\
        *   In the first panel, the shading is concentrated on the \textbf{left} side. \\
        *   In the second panel, the shading moves to the \textbf{top}. \\
        *   In the third panel, the shading moves to the \textbf{bottom-right}. \\
    *   This indicates a general \textbf{clockwise rotation} of the shaded pattern around the shape.

2.  \textbf{Analyze the Second Group of Figures:} \\
    *   This group features a standard five-pointed star where pairs of adjacent "arms" or points are shaded. \\
    *   Let's trace the movement of these shaded pairs: \\
        *   \textbf{Panel 1:} The \textbf{Top} and \textbf{Top-Right} points are shaded. \\
        *   \textbf{Panel 2:} The shading moves one position clockwise; now the \textbf{Top-Right} and \textbf{Bottom-Right} points are shaded. \\
    *   To maintain a consistent regularity, the next figure should continue this pattern. \\
    *   \textbf{Panel 3 (Expected):} Moving one more position clockwise from Panel 2, the \textbf{Bottom-Right} and \textbf{Bottom-Left} points should be shaded.

3.  \textbf{Evaluate the Options:} \\
    *   \textbf{A:} Top and Bottom-Right points are shaded. \\
    *   \textbf{B:} Bottom-Right and Bottom-Left points are shaded. \\
    *   \textbf{C:} Bottom-Left and Top-Left points are shaded. \\
    *   \textbf{D:} Top-Left and Top points are shaded.

\textbf{Correct Answer: C}
\end{wrongbox}

\vspace{5pt}

\begin{wrongbox}{Vanilla Qwen3-VL-8B \xmark}
  \small
To solve this problem, we analyze the patterns in both groups of figures.  \\
\textbf{First Group Analysis} (Left Side: Three Figures)  \\
Each figure is a star with one black triangular segment. Observing the sequence:  \\- The black segment **rotates clockwise** by a consistent angle between each figure. From the first to the second figure, and then to the third, the black segment moves one position clockwise around the star’s points. This establishes a **clockwise rotation pattern** for the black segment.  \\ \textbf{Second Group Analysis} (Right Side: Two Figures + ?)  \\The first two figures in the second group are stars with a white inner star and one black triangular segment. We apply the same rotational logic as the first group:  \\- The black segment in the second group must also follow a **clockwise rotation** pattern. The first figure has the black segment at a specific position, the second figure shows it rotated clockwise by one position, so the third figure (the answer) should show the black segment rotated **one more position clockwise** from the second figure.  \\ \textbf{Option Evaluation}  \\- **A**: The black segment is in a position that does not align with a clockwise rotation from the second figure in the second group.  \\- **B**: The black segment position is inconsistent with the established clockwise rotation pattern.  \\- **C**: The black segment is incorrectly placed relative to the expected clockwise rotation.  \\- **D**: The black segment is positioned exactly where a clockwise rotation from the second figure in the second group would place it, matching the pattern observed in the first group.  \\Thus, the correct answer is **D**.
\end{wrongbox}

\vspace{5pt}

\begin{rightbox}{Qwen3-VL-8B+MMInduction (Ours) \cmark}
  \scriptsize
  \setlength{\columnsep}{12pt}
  \setlength{\parskip}{2pt}
  \linespread{0.95}\selectfont
  <|begin\_of\_target\_analysis|>\\
{[Core Task]}: Determine which option logically completes the sequence of star shapes.\\
{[Key Visuals]}: The sequence consists of two groups of three identical star shapes. Each star has five points, with alternating black and white triangles at each point. The black triangles are consistently oriented in the same direction across the sequence.\\
{[Required Logic]}: The task requires identifying a consistent geometric transformation or pattern across the entire sequence, rather than focusing solely on the orientation of individual triangles within a single star.\\
<|end\_of\_target\_analysis|>

<|begin\_of\_reference\_analysis|>\\
<|begin\_of\_case\_1\_analysis|>\\
{[Elements]}: A sequence of six identical star shapes with alternating black and white triangles. The question asks to find the missing shape.\\
{[Solution Logic]}: The pattern involves a systematic rotation of the star around its center. Each subsequent figure is obtained by rotating the previous one by a fixed angle.\\
{[Comparison]}: Highly relevant. Both involve a sequence of identical star shapes where the orientation of the triangles changes systematically.\\
<|begin\_of\_helpful\_judgment|>helpful<|end\_of\_helpful\_judgment|>\\
<|end\_of\_case\_1\_analysis|>

<|begin\_of\_case\_2\_analysis|>\\
{[Elements]}: A sequence of diamond-shaped figures containing various geometric shapes (circles, triangles, stars). The question asks to find the missing figure.\\
{[Solution Logic]}: The pattern relies on the systematic movement and rotation of multiple distinct shapes within the diamond.\\
{[Comparison]}: Relevant. It reinforces the concept of tracking the relative positions and orientations of multiple elements within a composite shape.\\
<|begin\_of\_helpful\_judgment|>helpful<|end\_of\_helpful\_judgment|>\\
<|end\_of\_case\_2\_analysis|>

<|begin\_of\_case\_3\_analysis|>\\
{[Elements]}: A sequence of star-like radial lines with small triangles at their ends. The question asks to find the missing figure.\\
{[Solution Logic]}: The pattern involves a cyclic rotation of the small triangles around the central axis.\\
{[Comparison]}: Relevant. It highlights the importance of tracking the orientation of specific elements (in this case, triangles) around a central axis.\\
<|begin\_of\_helpful\_judgment|>helpful<|end\_of\_helpful\_judgment|>\\
<|end\_of\_case\_3\_analysis|>\\
<|end\_of\_reference\_analysis|>

<|begin\_of\_induction|>\\
{[Helpful Cases Cited]}: <|case\_1|>, <|case\_2|>, <|case\_3|>\\
{[Extracted Rule]}: In sequences involving composite shapes with multiple internal elements, the correct solution is found by identifying a consistent geometric transformation applied to the entire shape, rather than merely focusing on the orientation of individual elements within the shape.\\
<|end\_of\_induction|>

<|begin\_of\_answer|>\\
{[Step-by-step Application]}: \\
1. Observe the first group of three stars. The black triangles are consistently oriented to the left.\\
2. Observe the second group of three stars. The black triangles are consistently oriented to the right.\\
3. Apply the extracted rule: The entire sequence is governed by a consistent geometric transformation applied to the entire star shape.\\
4. Evaluate the options: Option A shows the black triangles oriented to the left, matching the first group. Option B shows the black triangles oriented to the right, matching the second group. Option C shows the black triangles oriented to the left, but the overall shape is rotated clockwise compared to the first group. Option D shows the black triangles oriented to the right, but the overall shape is rotated counterclockwise compared to the second group.\\
5. Conclusion: Option B is the only option that perfectly matches the geometric transformation required to complete the sequence.\\
{[Final Result]}: B\\
<|end\_of\_answer|>
\end{rightbox}

\vspace{4pt}
{\setlength{\abovecaptionskip}{6pt}%
\captionof{figure}{Case study on the MMIQ dataset. Vanilla Gemini-3.1-pro and Qwen3-VL fail to reach the correct answer, while our \textbf{MMInduction} induces the underlying rule from retrieved demonstrations and performs step-by-step reasoning to reach the correct answer.}%
\label{fig:case_study_mmiq_a_only}}

\FloatBarrier

\newpage
\FloatBarrier
\par\addvspace{4pt}

\noindent
\begin{headerbox}
\centering
\textbf{Dataset:} MDK12 \quad$|$\quad
\textbf{Task:} STEM Reasoning \quad$|$\quad
\textbf{Case ID:} \#2878
\end{headerbox}

\vspace{5pt}

\noindent
\begin{minipage}[t]{0.48\linewidth}
\vspace{0pt}
\tcbset{equal height group=casestudy-all_wrong-MDK12-2878}
  \begin{querybox}
    \centering
    \includegraphics[width=0.72\linewidth]{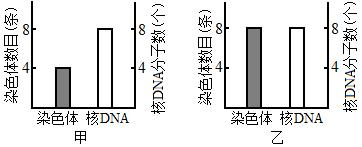}\\[4pt]
    \raggedright\small
    \textbf{Q:} The male fruit fly (2n=8) has the genotype AaBb, with genes A and B located on the same autosome. During meiosis of a spermatogonium in this male fruit fly, a crossover occurs between the non-sister chromatids of homologous chromosomes, producing a sperm with the genotype Ab. The number of chromosomes and nuclear DNA molecules in this spermatogonium during two different stages of meiosis is shown in the diagram below. Which of the following statements is incorrect? (\quad) \\
A. The number of chromatids in cells during both stages A and B is 8. \\
B. The cell during stage B may contain either 0 or 2 X chromosomes. \\
C. The genotype of a sperm from another secondary spermatocyte can be either ab or aB. \\
D. If this male fruit fly is test-crossed with a female fruit fly with the genotype aabb, and the offspring segregation ratio is 45:5:5:45\par\vspace{2pt}
    \textbf{GT Answer:} A, C
  \end{querybox}
\end{minipage}\hfill
\begin{minipage}[t]{0.50\linewidth}
\vspace{0pt}
\tcbset{equal height group=casestudy-all_wrong-MDK12-2878}
  \begin{demobox}
    \small
    \begin{minipage}[c]{0.26\linewidth}\centering
      \includegraphics[width=\linewidth]{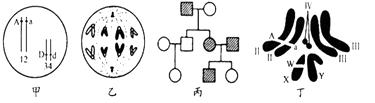}
    \end{minipage}\hfill
    \begin{minipage}[c]{0.70\linewidth}\raggedright
      \textbf{Q1:} The incorrect description of the biological significance indicated by the following figures is (\quad)

A. The probability of genotype Aadd individuals produced from self-crossing in Figure A is 1/6.

B. The cell in Figure B must be in the anaphase of mitosis, and the normal somatic cell chromosome number of this organism is 4.

C. In the family pedigree of Figure C, there are significantly more male patients than female patients, and the disease is most likely an X-linked recessive disorder.

D. Figure D shows the chromosomal composition of a specific fruit fly, with gamete genotypes being AX\textsuperscript{W}, aX\textsuperscript{W}, AY, and aY.\par
      \textbf{A1:} A, C
    \end{minipage}
    \par\vspace{3pt}\hrule\vspace{3pt}
    \begin{minipage}[c]{0.26\linewidth}\centering
      \includegraphics[width=\linewidth]{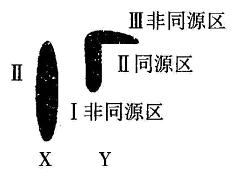}
    \end{minipage}\hfill
    \begin{minipage}[c]{0.70\linewidth}\raggedright
      \textbf{Q2:} If the diagram of partial chromosomes in a male cell is shown, the following statement is correct: 

A. A pathogenic gene is located in section II, and its inheritance follows the law of segregation. \\
B. A gene in section II on the X chromosome may have an allele on the Y chromosome. \\
C. Assuming the presence of an allele on the Y chromosome for a gene on the X chromosome, then this gene may have arisen from a mutation on the X chromosome. \\
D. If a pair of genes controlling a trait are located in section II, then this person has two genotypes.\par
      \textbf{A2:} A, B
    \end{minipage}
  \end{demobox}
\end{minipage}

\vspace{8pt}

\tcbset{breakable, enforce breakable}

\begin{wrongbox}{Vanilla Gemini-3.1-pro \xmark}
  \small
  To determine which statement is **incorrect**, let's analyze the scenario and each option carefully.

---

\#\#\# Given: \\
- Male fruit fly, 2n = 8 → so in somatic cells: 8 chromosomes. \\
- Genotype: AaBb, with **A and B on the same autosome** → meaning they are **linked** on one homologous pair. \\
- During meiosis of a **spermatogonium**, a **crossover** occurs between non-sister chromatids of homologous chromosomes → produces a sperm with genotype **Ab**. \\
- We're given two bar charts (Jia and Yi) showing chromosome number and nuclear DNA molecule count at two stages.

Let's interpret the diagrams:

\#\#\#\# Diagram Jia (Stage A): \\
- Chromosome number = 4 \\
- Nuclear DNA molecules = 8 \\
→ This suggests **haploid chromosome number but duplicated DNA** → typical of **secondary spermatocyte after meiosis I**, before meiosis II division. Each chromosome has 2 chromatids → so 4 chromosomes × 2 = 8 DNA molecules.

\#\#\#\# Diagram Yi (Stage B): \\
- Chromosome number = 8 \\
- Nuclear DNA molecules = 8 \\
→ This suggests **diploid chromosome number, unduplicated DNA** → typical of **sperm cells after meiosis II**? Wait — no! After meiosis II, chromosomes should be 4 (haploid), not 8.

Wait — this is confusing. Let's think again.

Actually, in meiosis:

- **Spermatogonium (before S phase)**: 8 chromosomes, 8 DNA \\
- **After S phase (primary spermatocyte)**: 8 chromosomes, 16 DNA (each chromosome has 2 chromatids) \\
- **After Meiosis I (secondary spermatocytes)**: 4 chromosomes, 8 DNA (each chromosome still has 2 chromatids) \\
- **After Meiosis II (sperm)**: 4 chromosomes, 4 DNA

Another possibility: Yi might represent **anaphase II**, where sister chromatids have separated, so temporarily 8 chromosomes (but still in one cell?) — but in anaphase II, the cell has 8 chromatids moving apart, but they are now called chromosomes — so yes, **in anaphase II**, chromosome number doubles temporarily to 8 (since sister chromatids separate), and DNA molecules = 8 (each chromosome now single-chromatid).

That makes sense! The question asks for which is incorrect, so \boxed{D} 

In the options, it's listed as D, and in the answer format, we box the letter.

So \boxed{D}
\end{wrongbox}

\vspace{5pt}

\begin{wrongbox}{Vanilla Qwen3-VL-8B \xmark}
  \small
  To determine the incorrect statements, let's analyze each option based on the principles of meiosis and the specific characteristics of Drosophila (fruit flies):

1.  **Analyze the diagrams (Stages A and B):** \\
    *   **Figure A (Jia):** The cell has 4 chromosomes and 8 nuclear DNA molecules. Since the normal somatic cell of a fruit fly has $2n = 8$ chromosomes, a cell with 4 chromosomes and 8 DNA molecules indicates that homologous chromosomes have separated, but sister chromatids have not. This corresponds to a secondary spermatocyte in **Prophase II or Metaphase II**. The number of chromatids here is 8. \\
    *   **Figure B (Yi):** The cell has 8 chromosomes and 8 nuclear DNA molecules. This temporary doubling of chromosomes (from 4 to 8) with an equal number of DNA molecules indicates that the centromeres have split and sister chromatids have separated. This corresponds to a secondary spermatocyte in **Anaphase II or Telophase II**. At this stage, because the sister chromatids have separated into independent chromosomes, the number of chromatids is **0**. \\
    *   **Conclusion for Option A:** The number of chromatids in stage B is 0, not 8. Therefore, statement A is **incorrect**.

2.  **Analyze the sex chromosomes (Option B):** \\
    *   A male fruit fly has sex chromosomes XY. During Meiosis I, homologous chromosomes (X and Y) separate into different secondary spermatocytes. \\
    *   Thus, a secondary spermatocyte will contain either one X or one Y chromosome.  \\
    *   In stage B (Anaphase II), the centromeres split. A cell that received the X chromosome will now have **2 X chromosomes**. A cell that received the Y chromosome will have 2 Y chromosomes and **0 X chromosomes**.  \\
    *   **Conclusion for Option B:** Statement B is correct.

3.  **Analyze the genotypes of sperms produced (Option C):** \\
    *   The primary spermatocyte has the genotype $AaBb$. A crossover occurred between non-sister chromatids of homologous chromosomes, producing 4 chromatids for this bivalent: $AB$, $Ab$, $aB$, and $ab$. \\
    *   During Meiosis I, the homologous centromeres separate.  \\
        *   If the centromere attached to the $A$ allele goes to one pole, that secondary spermatocyte will produce sperms with genotypes $AB$ and $Ab$.  \\
        *   The other secondary spermatocyte will receive the centromere attached to the $a$ allele and will produce sperms with genotypes $aB$ and $ab$. \\
    *   Alternatively, depending on the relative positions of the genes and the centromere, the other secondary spermatocyte could produce $AB$ and $aB$. In either valid scenario, it is entirely possible for the other secondary spermatocyte to produce a sperm with genotype $ab$ or $aB$. \\
    *   **Conclusion for Option C:** Statement C is correct.

4.  **Analyze the test-cross of a male fruit fly (Option D):** \\
    *   A fundamental rule in genetics is that **male Drosophila do not normally undergo crossing over** during meiosis; their linked genes exhibit complete linkage. \\
    *   Although the problem states that a rare crossover occurred in *one specific spermatogonium*, the vast majority of spermatogonia will undergo normal meiosis without crossing over.  \\
    *   Therefore, if this male fruit fly is test-crossed with an $aabb$ female, the offspring will overwhelmingly display the parental phenotypes in a roughly **1:1** ratio (i.e., 50:50:0:0). It will not yield a 45:5:5:45 ratio, which would imply a regular 10\% recombination frequency across all gametes. \\
    *   **Conclusion for Option D:** Statement D is **incorrect**.

Therefore, the incorrect statements are A and D.

Answer: A, D
\end{wrongbox}

\vspace{2pt}

\begin{wrongbox}{MMInduction (Ours) \xmark}
  \scriptsize
  \setlength{\columnsep}{12pt}
  \setlength{\parskip}{2pt}
  \linespread{0.95}\selectfont
  <|begin\_of\_target\_analysis|>\\
{[Core Task]}: Determine which statement about the meiotic process and resulting sperm genotypes is incorrect based on the provided bar graphs showing chromosome and DNA numbers at two stages of meiosis.\\
{[Key Visuals]}: Two bar graphs (Jia and Yi). Graph Jia shows a cell with 4 chromosomes and 8 DNA molecules (indicating 4 chromatids), while Graph Yi shows a cell with 8 chromosomes and 8 DNA molecules (indicating 8 chromatids).\\
{[Required Logic]}: Apply the principles of meiosis, including the behavior of homologous chromosomes, sister chromatids, and sex chromosomes, to deduce the exact stage of meiosis and the possible genotypes of the sperm.\\
<|end\_of\_target\_analysis|>

<|begin\_of\_reference\_analysis|>\\
<|begin\_of\_case\_1\_analysis|>\\
{[Elements]}: The image contains multiple biological figures (chromosomes, mitosis, pedigrees, karyotype). The text asks to identify incorrect biological descriptions.\\
{[Solution Logic]}: The solution requires interpreting genetic diagrams and pedigrees to determine the probability of genotypes, chromosome counts, and inheritance patterns.\\
{[Comparison]}: While it involves genetics, it focuses on pedigree analysis and chromosome counting rather than the detailed mechanics of meiosis or the specific behavior of sex chromosomes.\\
<|begin\_of\_helpful\_judgment|>unhelpful<|end\_of\_helpful\_judgment|>\\
<|end\_of\_case\_1\_analysis|>

<|begin\_of\_case\_2\_analysis|>\\
{[Elements]}: The image shows a simplified diagram of the X and Y chromosomes with three regions (I, II, III). The text asks to judge statements about the inheritance of genes located in these regions.\\
{[Solution Logic]}: The solution relies on understanding the structure of sex chromosomes and the specific inheritance patterns of genes located in the non-homologous regions (II).\\
{[Comparison]}: This case is highly relevant because it specifically addresses the inheritance and behavior of genes on sex chromosomes, which is a critical component of the target problem involving a male fruit fly.\\
<|begin\_of\_helpful\_judgment|>helpful<|end\_of\_helpful\_judgment|>\\
<|end\_of\_case\_2\_analysis|>\\
<|end\_of\_reference\_analysis|>

<|begin\_of\_induction|>\\
{[Helpful Cases Cited]}: <|case\_2|>\\
{[Extracted Rule]}: In a male fruit fly (XY), the X and Y chromosomes are structurally distinct. Genes located on the X chromosome are inherited according to the law of independent assortment. However, genes located on the Y chromosome are only passed down to sons. If a crossover occurs between non-sister chromatids of homologous chromosomes, it can result in the exchange of genetic material between the X and Y chromosomes, leading to the formation of sex-reversed gametes (e.g., X or Y).\\
<|end\_of\_induction|>

<|begin\_of\_answer|>\\
{[Step-by-step Application]}:\\
1. **Analyze the Bar Graphs**: The first graph (Jia) shows 4 chromosomes and 8 DNA molecules, indicating that each chromosome consists of 2 sister chromatids. This corresponds to the prophase I stage of meiosis when homologous chromosomes pair up and undergo crossing over. The second graph (Yi) shows 8 chromosomes and 8 DNA molecules, indicating that each chromosome now consists of 4 sister chromatids. This corresponds to metaphase I or metaphase II of meiosis.\\
2. **Determine the Stage**: Since the number of chromosomes doubles from 4 to 8, the cell must be undergoing metaphase I . At this stage, the cell contains 4 chromosomes (2 pairs of homologues) and 4 X chromosomes (2 pairs of homologues). The cell cannot be in metaphase II  because M II would show 4 chromosomes (2 pairs of sister chromatids) and 4 DNA molecules.\\
3. **Evaluate Option A**: Both stages (prophase I and metaphase I) involve 4 pairs of sister chromatids, so the number of chromatids is always 8. Statement A is correct.\\
4. **Evaluate Option B**: During metaphase I, the cell may contain either 0 or 2 X chromosomes depending on whether the homologous X chromosomes are aligned on opposite poles. Statement B is correct.\\
5. **Evaluate Option C**: The sperm genotype is Ab. Since the male fruit fly has the genotype AaBb, the other possible sperm genotypes are ab and aB. Therefore, the genotype of a sperm from another secondary spermatocyte can indeed be either ab or aB. Statement C is correct.\\
6. **Evaluate Option D**: The given offspring segregation ratio (45:5:5:45) is characteristic of a test cross between a heterozygous recessive individual (aaBb) and a homozygous recessive individual (aabb). This indicates that the male fruit fly is heterozygous for the B locus (AaBb). However, the question asks which statement is *incorrect*. Since options A, B, and C are all correct, option D is the only incorrect statement.\\
{[Final Result]}: D\\
<|end\_of\_answer|>
\end{wrongbox}

\vspace{4pt}
{\setlength{\abovecaptionskip}{6pt}%
\captionof{figure}{Failure case on the MDK12 dataset where all three methods (Vanilla Gemini-3.1-pro, Vanilla Qwen3-VL, and our \textbf{MMInduction}) fail to reach the correct answer.}%
\label{fig:case_study_mdk12_all_wrong}}

\FloatBarrier

\newpage
\FloatBarrier
\par\addvspace{4pt}

\noindent
\begin{headerbox}
\centering
\textbf{Dataset:} MAMI \quad$|$\quad
\textbf{Task:} Multimodal Knowledge Reasoning \quad$|$\quad
\textbf{Case ID:} \#41d3aaac-4a0a-432f-93d8-dd6df69e88d7
\end{headerbox}

\vspace{5pt}

\noindent
\begin{minipage}[t]{0.48\linewidth}
\vspace{0pt}
\tcbset{equal height group=casestudy-all_correct-MAMI-41d3aaac-4a0a-432f-93d8-dd6df69e88d7}
  \begin{querybox}
    \centering
    \includegraphics[width=0.72\linewidth]{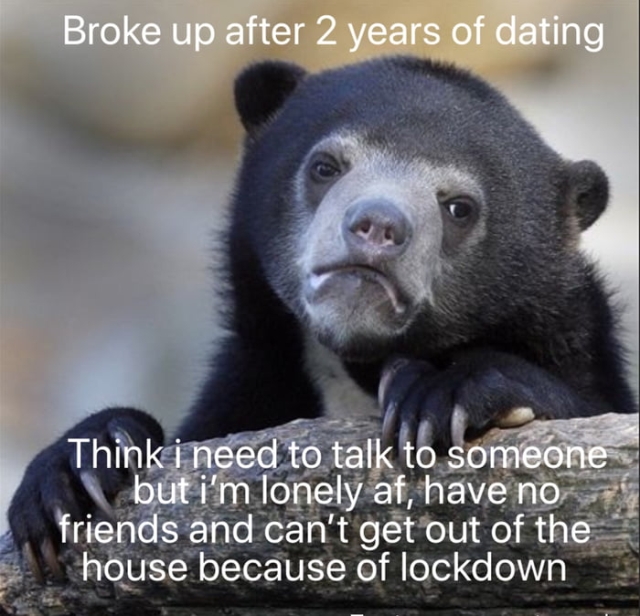}\\[4pt]
    \raggedright\small
    \textbf{Q:} Determine whether this meme is misogynistic.\par\vspace{2pt}
    \textbf{GT Answer:} False
  \end{querybox}
\end{minipage}\hfill
\begin{minipage}[t]{0.50\linewidth}
\vspace{0pt}
\tcbset{equal height group=casestudy-all_correct-MAMI-41d3aaac-4a0a-432f-93d8-dd6df69e88d7}
  \begin{demobox}
    \small
    \begin{minipage}[c]{0.26\linewidth}\centering
      \includegraphics[width=\linewidth]{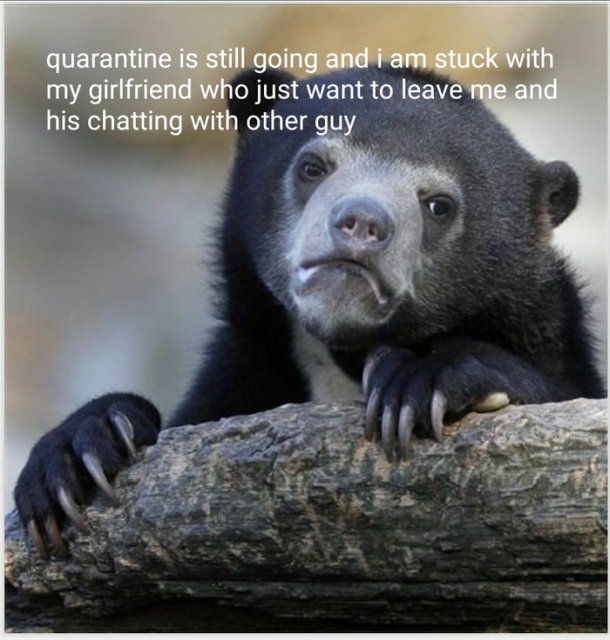}
    \end{minipage}\hfill
    \begin{minipage}[c]{0.70\linewidth}\raggedright
      \textbf{Q1:} Determine whether this meme is misogynistic.\par
      \textbf{A1:} False
    \end{minipage}
    \par\vspace{3pt}\hrule\vspace{3pt}
    \begin{minipage}[c]{0.26\linewidth}\centering
      \includegraphics[width=\linewidth]{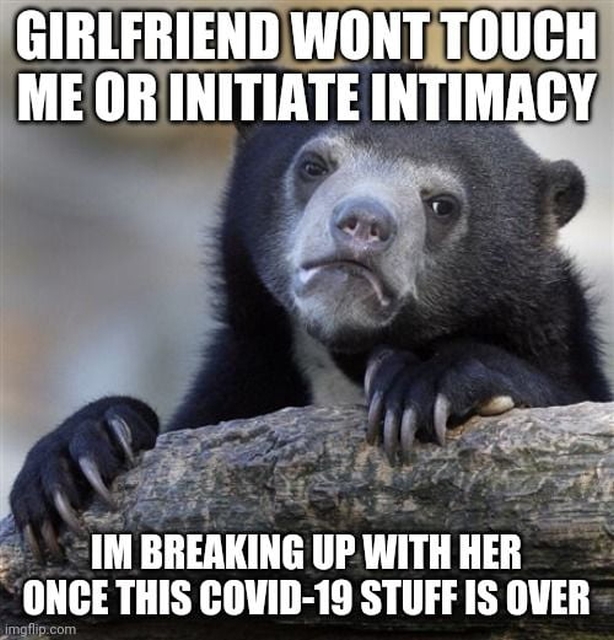}
    \end{minipage}\hfill
    \begin{minipage}[c]{0.70\linewidth}\raggedright
      \textbf{Q2:} Determine whether this meme is misogynistic.\par
      \textbf{A2:} False
    \end{minipage}
    \par\vspace{3pt}\hrule\vspace{3pt}
    \begin{minipage}[c]{0.26\linewidth}\centering
      \includegraphics[width=\linewidth]{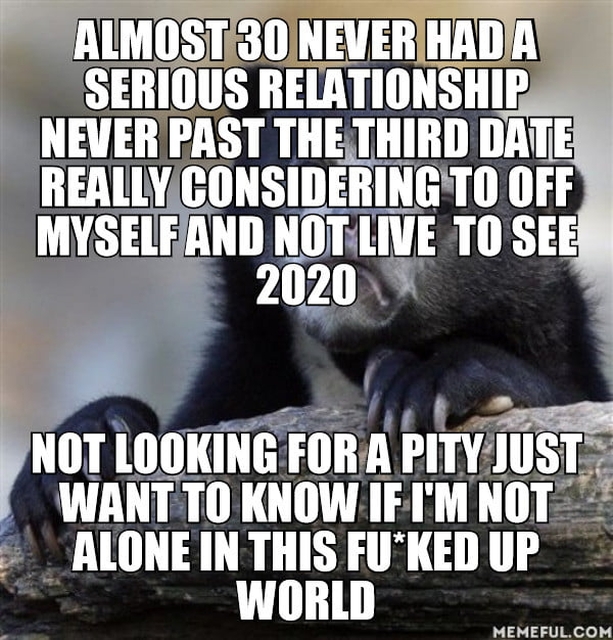}
    \end{minipage}\hfill
    \begin{minipage}[c]{0.70\linewidth}\raggedright
      \textbf{Q3:} Determine whether this meme is misogynistic.\par
      \textbf{A3:} False
    \end{minipage}
  \end{demobox}
\end{minipage}

\vspace{8pt}

\tcbset{breakable, enforce breakable}

\begin{rightbox}{MMInduction (Ours) \cmark}
  \scriptsize
  \setlength{\columnsep}{12pt}
  \setlength{\parskip}{2pt}
  \linespread{0.95}\selectfont
  <|begin\_of\_target\_analysis|>\\
{[Core Task]}: Determine whether the target meme is misogynistic.\\
{[Key Visuals]}: A sad bear meme with text saying “Broke up after 2 years of dating” and “Think i need to talk to someone but i’m lonely af, have no friends and can’t get out of the house because of lockdown.”\\
{[Required Logic]}: Assess whether the meme expresses hatred, contempt, stereotyping, or derogatory content toward women. If it is just about loneliness, breakup, or social isolation without attacking women, then it is not misogynistic.\\
<|end\_of\_target\_analysis|>

<|begin\_of\_reference\_analysis|>\\
<|begin\_of\_case\_1\_analysis|>\\
{[Elements]}: The meme shows a bear with text about quarantine, being stuck with a girlfriend who wants to leave, and chatting with another guy. The question asks whether it is misogynistic.\\
{[Solution Logic]}: The meme discusses relationship frustration and social isolation, not hostility toward women as a group. Therefore it is not misogynistic.\\
{[Comparison]}: This is highly similar to the target: both involve a breakup/relationship context and express feelings of loneliness/social isolation rather than anti-women sentiment.\\
<|begin\_of\_helpful\_judgment|>helpful<|end\_of\_helpful\_judgment|>\\
<|end\_of\_case\_1\_analysis|>

<|begin\_of\_case\_2\_analysis|>\\
{[Elements]}: The meme says the girlfriend won’t touch or initiate intimacy, and the speaker says he is breaking up with her once COVID stuff is over. The question asks whether it is misogynistic.\\
{[Solution Logic]}: Even though it mentions a girlfriend, the meme’s main message is about personal dissatisfaction and breakup, not contempt for women. So it is not misogynistic.\\
{[Comparison]}: Very relevant. Like the target, it centers on a breakup and emotional distress rather than gender-based hostility.\\
<|begin\_of\_helpful\_judgment|>helpful<|end\_of\_helpful\_judgment|>\\
<|end\_of\_case\_2\_analysis|>

<|begin\_of\_case\_3\_analysis|>\\
{[Elements]}: The meme says the speaker has never had a serious relationship, is considering suicide, and asks if others feel alone in 2020. The question asks whether it is misogynistic.\\
{[Solution Logic]}: The meme is about depression, isolation, and self-doubt, not about women or gender-based prejudice. Thus it is not misogynistic.\\
{[Comparison]}: Relevant because the target also conveys loneliness and isolation, not misogyny.\\
<|begin\_of\_helpful\_judgment|>helpful<|end\_of\_helpful\_judgment|>\\
<|end\_of\_case\_3\_analysis|>\\
<|end\_of\_reference\_analysis|>

<|begin\_of\_induction|>\\
{[Helpful Cases Cited]}: <|case\_1|>, <|case\_2|>, <|case\_3|>\\
{[Extracted Rule]}: A meme should be classified as misogynistic only if it contains clear hostility, contempt, derogation, or negative stereotyping directed at women as a class. If the meme instead expresses personal sadness, loneliness, relationship frustration, or social isolation without targeting women, then it is not misogynistic.\\
<|end\_of\_induction|>

<|begin\_of\_answer|>\\
{[Step-by-step Application]}: The target meme describes feeling lonely, having no friends, and being unable to leave the house due to lockdown. It also mentions breaking up after dating for two years. These are personal emotional statements about isolation and relationship issues. There is no statement about women, no insult toward women, and no gender-based stereotype. Applying the extracted rule, the meme is not misogynistic.\\
{[Final Result]}: False\\
<|end\_of\_answer|>
\end{rightbox}

\vspace{4pt}
{\setlength{\abovecaptionskip}{6pt}%
\captionof{figure}{Case study on the MAMI dataset where all three methods reach the correct answer. Our \textbf{MMInduction} induces the underlying rule from retrieved demonstrations and performs step-by-step reasoning.}%
\label{fig:case_study_mami_all_correct}}

\FloatBarrier

As illustrated in the VQA example Figure \ref{fig:case_study_vqa}, tasks relying on surface-level visual similarity are generally well-handled by both vanilla VLMs and our MMInduction framework. Vanilla models can quickly grasp the task intent from the retrieved cases and output the correct answer based on visual matching. However, MMInduction further enhances this process by explicitly articulating the visual grounding rule, such as locating the specific subject and inspecting its attributes, thereby providing a transparent and interpretable reasoning trajectory even for straightforward perception tasks.

Figure \ref{fig:case_study_mmiq_a_only} present complex logical reasoning cases from the MMIQ dataset, highlighting the critical failure of vanilla models. In these instances, vanilla VLMs either hallucinate incorrect patterns or fail to generalize spatial transformation rules across the sequence, leading to erroneous conclusions. Conversely, MMInduction successfully navigates these challenges. By systematically analyzing the reference cases, the model accurately induces the underlying geometric transformation rules (e.g., consistent clockwise rotation) and rigorously applies them to the target query, demonstrating a profound improvement in cross-case logical induction.

Figure \ref{fig:case_study_mdk12_all_wrong} illustrates a challenging STEM question in genetics, requiring the model to reason about meiosis, crossover, and sperm genotype determination in fruit flies. Both vanilla Gemini‑3.1‑pro and vanilla Qwen3‑VL‑8B fail to reach the correct answer (A and C), producing logically inconsistent or incomplete responses. Our MMInduction framework successfully extracts the relevant genetic principles from the retrieved reference cases—such as the behavior of sex chromosomes and the impact of crossing over on gamete genotypes—and follows the inductive‑deductive CoT to structure its reasoning. However, MMInduction also fails to output the correct final answer. As shown in the example, the model correctly identifies parts of the problem but makes an error in evaluating one of the multi‑select options (e.g., misjudging statement D), leading to an incomplete solution. This failure reveals that while MMInduction substantially improves cross‑case rule induction and reasoning structure, the underlying VLM’s lack of precise domain‑specific knowledge and step‑wise logical verification still imposes an upper bound on performance in rigorous STEM tasks. Future work should address this through stronger base models or tighter integration with external knowledge and verification tools.

Figure \ref{fig:case_study_mami_all_correct} showcases the effectiveness of MMInduction in nuanced social intelligence tasks, such as identifying misogynistic memes in the MAMI dataset. Vanilla models often struggle with implicit semantics and may misinterpret personal complaints as targeted hate speech. Through the Inductive CoT, our framework effectively synthesizes a clear classification rule from the retrieved examples—distinguishing between general emotional distress and explicit gender-based hostility. By applying this induced rule, MMInduction accurately evaluates the target meme's text and visual elements, successfully avoiding false positives and demonstrating robust multimodal semantic understanding.


\section{Limitations}
\label{app:limitation}

Despite the  empirical results achieved by MMInduction, several inherent limitations remain.

First, the inductive-deductive Chain-of-Thought significantly increases the length of generated sequences. As shown in Table \ref{tab:hyper}, extending the number of in-context cases to four or five shots leads to a pronounced decline in the success rate (from 91.75 at 3-shot to 69.48 at 5-shot). The model tends to over-analyze cases, producing excessively long responses that frequently exceed the context window, causing early termination or reasoning failures. This indicates that the base model’s capacity for stable long-form reasoning remains a bottleneck, and our framework does not fundamentally resolve the length-vs-depth trade-off inherent to multi-shot ICL.

Second, while MMInduction substantially improves logical and visual reasoning, it does not eliminate all failure modes. As illustrated in the MDK12 case study (Figure \ref{fig:case_study_mdk12_all_wrong}), even our method fails on complex STEM problems that require multi-step algebraic manipulation. The model successfully extracts the relevant geometric formulas from reference cases but falters during the numerical calculation phase, revealing that the underlying VLM’s mathematical computation capability imposes an upper bound on performance in rigorous quantitative domains. Addressing this would require either stronger base models or tighter integration of external computation tools.

Third, our framework relies on similarity-based retrieval to select in-context demonstrations. Although we show that random retrieval degrades performance (Table \ref{tab:recall}), the retrieval quality still depends on the embedding model used. Qwen3VL-Emb provides stable matching, but the approach may be less effective when the knowledge base lacks sufficient semantic diversity or when the target task involves highly novel patterns not well represented in the retrieval corpus. The generalizability of MMInduction to truly zero-retrieval or adversarially constructed few-shot settings remains to be fully verified.

Finally, our training pipeline (SFT + RLVR) distills reasoning chains from proprietary models (Gemini-3.1-pro) and uses them as supervision for open-source VLMs. While this boosts performance, it inherits any latent biases or reasoning flaws present in the teacher model. Moreover, the composite reward function, though effective, requires careful tuning of hyper-parameters, and the DAPO-based optimization is computationally non-trivial, limiting accessibility for researchers with modest resources.

We view these limitations as opportunities for future work, including length-adaptive CoT pruning, integration with external verifiers for numerical reasoning, and more robust retrieval-free induction mechanisms.

\section{Social Impact}
\label{app:impact}

\paragraph{\textbf{Positive Societal Impacts.}} Our work aims to enhance the inductive reasoning capability of multimodal large language models, which has direct societal benefits. By enabling VLMs to genuinely extract generalizable rules from in-context demonstrations rather than relying on shallow pattern matching, MMInduction improves model performance on complex tasks such as logical reasoning, STEM question answering, and harmful content detection (e.g., misogynistic meme identification on the MAMI dataset). This advancement can support educational tools by providing interpretable step-by-step reasoning for students, assist content moderators in identifying nuanced forms of online harm, and improve the reliability of vision-language systems in high-stakes applications such as scientific diagram analysis or medical visual question answering. The explicit inductive-deductive reasoning trace also enhances model transparency, allowing users to inspect how a conclusion was reached, which is a critical requirement for trustworthy AI deployment.

\paragraph{\textbf{Negative Societal Impacts.}} Despite these benefits, MMInduction may introduce risks if deployed without appropriate safeguards. First, the model may still produce incorrect answers accompanied by seemingly coherent but flawed reasoning chains (the “inductive correctness gap” that this work partially addresses but does not fully eliminate). Such errors could mislead users who over-trust the model’s explanatory outputs, particularly in educational or professional decision-making contexts. Second, improved multimodal ICL capabilities could be misused to generate more persuasive disinformation or to automate the creation of adversarial examples that exploit the model’s inductive biases. Third, our framework relies on similarity-based retrieval of in-context examples; if the retrieval database contains biased or harmful content, the model may inadvertently propagate or amplify those biases. Fourth, the computational cost of SFT + RLVR training (requiring 8×A100 GPUs) may limit access to well-resourced institutions, exacerbating existing inequalities in AI research and development. We strongly recommend that any deployment of MMInduction include human oversight, robustness testing against adversarial inputs, and explicit confidence or uncertainty indicators attached to the model’s reasoning traces.

\section{LLM Usage}
\label{app:llm}

This research extensively utilized large language models (LLMs) as integral components of the proposed method and evaluation pipeline: Gemini-3.1-pro served as the teacher model to generate high-quality reasoning chains for supervised fine-tuning, with the synthesized data subsequently verified via majority voting among GPT-5.4, Claude-4.6-Sonnet, and Seed-1.8-pro; GPT-5.4-mini, Gemini-3.0-flash, and Claude-4.5-haiku were employed as LLM judges for answer accuracy and induction accuracy evaluation, while the RLVR phase leveraged the DAPO algorithm for policy optimization. Beyond these method-specific uses, LLMs also assisted in the broader research process, including early-stage idea conceptualization (e.g., formulating the “inductive correctness gap” hypothesis), literature survey (identifying and summarizing prior work on multimodal ICL, visual token pruning, and attention mechanisms), understanding technical concepts (e.g., DAPO, RLVR, similarity-guided pruning), and refining the paper’s writing for grammar, clarity, and conciseness. All scientific claims, experimental results, and final wording remain the sole responsibility of the authors, and all LLM usage complied with the respective providers’ terms of service without exposure of any personally identifiable or proprietary data.

\newpage
\section*{NeurIPS Paper Checklist}

The checklist is designed to encourage best practices for responsible machine learning research, addressing issues of reproducibility, transparency, research ethics, and societal impact. Do not remove the checklist: {\bf The papers not including the checklist will be desk rejected.} The checklist should follow the references and follow the (optional) supplemental material.  The checklist does NOT count towards the page
limit. 

Please read the checklist guidelines carefully for information on how to answer these questions. For each question in the checklist:
\begin{itemize}
    \item You should answer \answerYes{}, \answerNo{}, or \answerNA{}.
    \item \answerNA{} means either that the question is Not Applicable for that particular paper or the relevant information is Not Available.
    \item Please provide a short (1--2 sentence) justification right after your answer (even for \answerNA). 
\end{itemize}

{\bf The checklist answers are an integral part of your paper submission.} They are visible to the reviewers, area chairs, senior area chairs, and ethics reviewers. You will also be asked to include it (after eventual revisions) with the final version of your paper, and its final version will be published with the paper.

The reviewers of your paper will be asked to use the checklist as one of the factors in their evaluation. While \answerYes{} is generally preferable to \answerNo{}, it is perfectly acceptable to answer \answerNo{} provided a proper justification is given (e.g., error bars are not reported because it would be too computationally expensive'' or ``we were unable to find the license for the dataset we used''). In general, answering \answerNo{} or \answerNA{} is not grounds for rejection. While the questions are phrased in a binary way, we acknowledge that the true answer is often more nuanced, so please just use your best judgment and write a justification to elaborate. All supporting evidence can appear either in the main paper or the supplemental material, provided in appendix. If you answer \answerYes{} to a question, in the justification please point to the section(s) where related material for the question can be found.

IMPORTANT, please:
\begin{itemize}
    \item {\bf Delete this instruction block, but keep the section heading ``NeurIPS Paper Checklist"},
    \item  {\bf Keep the checklist subsection headings, questions/answers and guidelines below.}
    \item {\bf Do not modify the questions and only use the provided macros for your answers}.
\end{itemize}


\begin{enumerate}

\item {\bf Claims}
    \item[] Question: Do the main claims made in the abstract and introduction accurately reflect the paper's contributions and scope?
    \item[] Answer: \answerYes{} 
    \item[] Justification: The abstract and introduction clearly state the three identified bottlenecks  and the three corresponding contributions.

\item {\bf Limitations}
    \item[] Question: Does the paper discuss the limitations of the work performed by the authors?
    \item[] Answer: \answerYes{} 
    \item[] Justification: Appendix Section \ref{app:limitation}

\item {\bf Theory assumptions and proofs}
    \item[] Question: For each theoretical result, does the paper provide the full set of assumptions and a complete (and correct) proof?
    \item[] Answer: \answerNA{} 
    \item[] Justification: The paper does not contain theoretical results (such as theorems, lemmas or formal proofs), and all the content is empirical research and algorithm design.

    \item {\bf Experimental result reproducibility}
    \item[] Question: Does the paper fully disclose all the information needed to reproduce the main experimental results of the paper to the extent that it affects the main claims and/or conclusions of the paper (regardless of whether the code and data are provided or not)?
    \item[] Answer: \answerYes{} 
    \item[] Justification: Appendix \ref{app:training} provides detailed experimental settings, including hardware configuration, training hyperparameters (Table \ref{tab:sft-rl-params}), data construction process, evaluation protocol, etc.; Appendix \ref{app:datasets} explains the splitting method of each dataset; Appendix \ref{app:prompt} presents all prompt templates. The paper also states that the code and data are included in the supplementary materials, sufficient to reproduce the main results.

\item {\bf Open access to data and code}
    \item[] Question: Does the paper provide open access to the data and code, with sufficient instructions to faithfully reproduce the main experimental results, as described in supplemental material?
    \item[] Answer: \answerYes{} 
    \item[] Justification: Code, data and partial results of the experiment are all provided in the supplementary materials.

\item {\bf Experimental setting/details}
    \item[] Question: Does the paper specify all the training and test details (e.g., data splits, hyperparameters, how they were chosen, type of optimizer) necessary to understand the results?
    \item[] Answer: \answerYes{} 
    \item[] Justification: Appendix \ref{app:training} provides detailed experimental settings, including hardware configuration, training hyperparameters (Table \ref{tab:sft-rl-params}), data construction process, evaluation protocol, etc.; Appendix \ref{app:datasets} explains the splitting method of each dataset; Appendix \ref{app:prompt} presents all prompt templates. The paper also states that the code and data are included in the supplementary materials, sufficient to reproduce the main results.

\item {\bf Experiment statistical significance}
    \item[] Question: Does the paper report error bars suitably and correctly defined or other appropriate information about the statistical significance of the experiments?
    \item[] Answer: \answerYes{} 
    \item[] Justification: All the experiments in the paper were repeated under three different random seeds, and the average accuracy rate was reported.

\item {\bf Experiments compute resources}
    \item[] Question: For each experiment, does the paper provide sufficient information on the computer resources (type of compute workers, memory, time of execution) needed to reproduce the experiments?
    \item[] Answer: \answerYes{} 
    \item[] Justification:  All experiments are conducted on 8×NVIDIA A100 GPUs with 80GB memory.
    
\item {\bf Code of ethics}
    \item[] Question: Does the research conducted in the paper conform, in every respect, with the NeurIPS Code of Ethics \url{https://neurips.cc/public/EthicsGuidelines}?
    \item[] Answer: \answerYes{} 
    \item[] Justification: The paper does not involve human subjects or sensitive data. All the datasets and models used are publicly available academic resources, and there are no obvious ethical violations.

\item {\bf Broader impacts}
    \item[] Question: Does the paper discuss both potential positive societal impacts and negative societal impacts of the work performed?
    \item[] Answer: \answerYes{} 
    \item[] Justification: Appendix Section \ref{app:impact}
    
\item {\bf Safeguards}
    \item[] Question: Does the paper describe safeguards that have been put in place for responsible release of data or models that have a high risk for misuse (e.g., pre-trained language models, image generators, or scraped datasets)?
    \item[] Answer: \answerNA{} 
    \item[] Justification: The paper did not introduce any new high-risk models or datasets. All the assets used were from publicly available resources and did not involve the generation of harmful content.

\item {\bf Licenses for existing assets}
    \item[] Question: Are the creators or original owners of assets (e.g., code, data, models), used in the paper, properly credited and are the license and terms of use explicitly mentioned and properly respected?
    \item[] Answer: \answerYes{} 
    \item[] Justification: All models and datasets used in this paper have been properly cited.

\item {\bf New assets}
    \item[] Question: Are new assets introduced in the paper well documented and is the documentation provided alongside the assets?
    \item[] Answer: \answerYes{} 
    \item[] Justification: We will provide the code as well as the README.md file to facilitate researcher communication.

\item {\bf Crowdsourcing and research with human subjects}
    \item[] Question: For crowdsourcing experiments and research with human subjects, does the paper include the full text of instructions given to participants and screenshots, if applicable, as well as details about compensation (if any)? 
    \item[] Answer: \answerNA{} 
    \item[] Justification: This paper did not involve any crowdsourcing or studies involving human subjects.

\item {\bf Institutional review board (IRB) approvals or equivalent for research with human subjects}
    \item[] Question: Does the paper describe potential risks incurred by study participants, whether such risks were disclosed to the subjects, and whether Institutional Review Board (IRB) approvals (or an equivalent approval/review based on the requirements of your country or institution) were obtained?
    \item[] Answer: \answerNA{} 
    \item[] Justification: This paper does not involve human subjects and does not require IRB approval.

\item {\bf Declaration of LLM usage}
    \item[] Question: Does the paper describe the usage of LLMs if it is an important, original, or non-standard component of the core methods in this research? Note that if the LLM is used only for writing, editing, or formatting purposes and does \emph{not} impact the core methodology, scientific rigor, or originality of the research, declaration is not required.
    \item[] Answer: \answerYes{} 
    \item[] Justification: Appendix Section \ref{app:llm}

\end{enumerate}

\end{document}